\documentclass[letterpaper]{article}
\usepackage[preprint]{aaai2027}
\usepackage[hyphens]{url} 
\usepackage{graphicx} 
\graphicspath{{./}{images/}}
\usepackage{natbib} 
\usepackage{caption} 
\usepackage{amsmath}
\usepackage{amsfonts}
\usepackage{booktabs}
\usepackage{multirow}
\usepackage[table]{xcolor}
\usepackage{hhline}
\usepackage{fontawesome5}
\usepackage{pifont}
\newcommand{\frozenmark}{\ding{100}}
\newcommand{\updatedmark}{\scalebox{0.82}{\faFire}}
\newcommand{\frozenprior}{\textsuperscript{\frozenmark}}
\newcommand{\updatedprior}{\textsuperscript{\updatedmark}}
\newcommand{\best}[1]{\leavevmode\rlap{\kern0.10pt#1}#1}
\usepackage{algorithm}
\usepackage{algorithmic}
\usepackage{newfloat}
\usepackage{listings}
\DeclareCaptionStyle{ruled}{labelfont=normalfont,labelsep=colon,strut=off} 
\floatstyle{ruled}
\newfloat{listing}{tb}{lst}{}
\floatname{listing}{Listing}
\title{Learning the Target Priors Before Image Translation: A Decoupled Training Paradigm for Cross-Modal Image Translation in Remote Sensing}
\author{
Keyan Hu\equalcontrib\textsuperscript{\rm 1},
Mingtao Wang\equalcontrib\textsuperscript{\rm 1},
Ziyu Zhou\textsuperscript{\rm 2},
Tiandong Shi\textsuperscript{\rm 1},
Haifeng Li\textsuperscript{\rm 1},
Ji Qi\textsuperscript{\rm 3}\corresponding,
Chao Tao\textsuperscript{\rm 1}\corresponding
}
\affiliations{}
\authornote{
\authornoteline{\textsuperscript{\rm 1}}{Central South University, Changsha, China.}
\authornoteline{\textsuperscript{\rm 2}}{Wuhan University, Wuhan, China.}
\authornoteline{\textsuperscript{\rm 3}}{Guangzhou University, Guangzhou, China.}
\authornoteline{}{Emails: phycheor@gmail.com; kingtaochao@csu.edu.cn.}
}
\begin{document}
\maketitle

\begin{abstract}
Cross-modal image translation in remote sensing must preserve source-observed content while matching the target-domain distribution. Existing methods jointly learn the target prior and cross-modal dependence from scarce paired data, overlooking a key asymmetry: only the latter intrinsically requires cross-modal correspondence. We formalize this distinction through conditional-score and denoising-risk analyses and propose Learning the Target Priors Before Image Translation (LTP-BIT), a prior-first paradigm that decouples the two learning tasks. LTP-BIT first learns a target-domain generative prior from large-scale unpaired imagery, then retains the pretrained backbone weights and learns source-conditioned control through P-DART, a parameter-efficient dual-stream architecture. Controlled experiments show that prior matching and scaling primarily improve target-domain realism, whereas instance fidelity relies more strongly on conditional adaptation. LTP-BIT achieves state-of-the-art performance across SAR-to-RGB and NIR-to-RGB benchmarks using only 9.81\% task-specific parameters. On QXS-SAROPT, it retains near-full-data instance fidelity with only 25\% of the paired samples.
\end{abstract}

\section{Introduction}

Cross-modal image translation in remote sensing predominantly learns $p(y\,\vert\,x)$ from limited image pairs~\citep{Zhang2022PDE,Huang2026GenAI}. Each pair provides both a plausible target image and its correspondence with the source observation, supporting target-domain realism and source-conditioned content fidelity, respectively. However, the task is inherently underdetermined because different sensors capture distinct physical properties of the same scene~\citep{Schmitt2016Fusion}. Source modalities such as SAR and NIR cannot fully determine visible-light-specific attributes, allowing multiple plausible target images for a single observation~\citep{Zhu2017Multimodal}. Resolving this ambiguity requires a target-domain prior to complete unobserved attributes and cross-modal dependence to constrain these completions. Crucially, only the latter intrinsically requires correspondence: the target marginal can be learned from unpaired target images. Conventional paired training entangles these two learning tasks, confining target-prior learning to the scale and coverage of the paired target set.

\begin{figure}[t]
  \centering
  \includegraphics[width=\columnwidth]{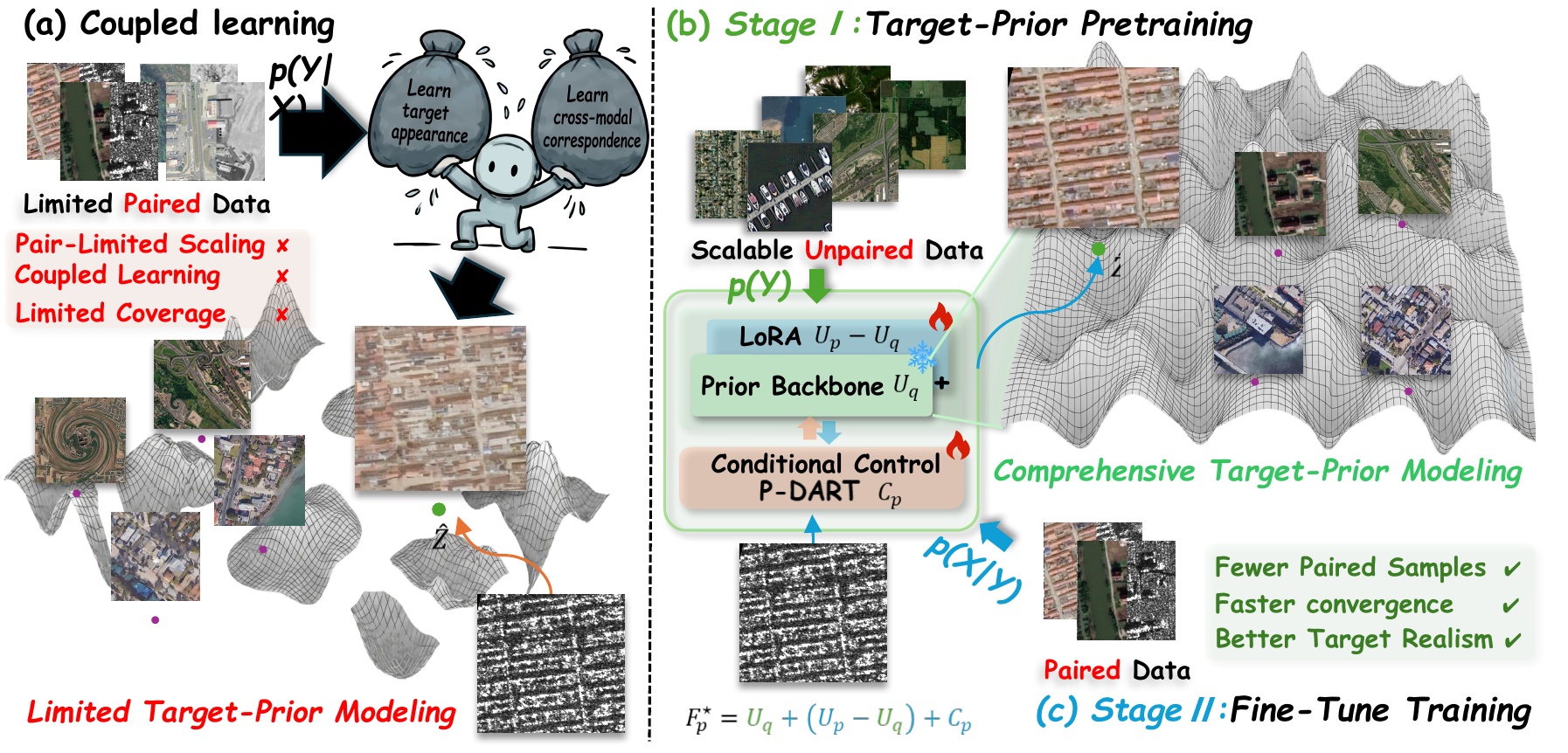}
  \caption{Overview of the prior-first learning paradigm in LTP-BIT for remote sensing cross-modal translation.}
  \label{fig:framework}
\end{figure}

Existing studies mitigate this ambiguity by constraining cross-modal mappings or introducing external knowledge. Task-specific methods impose generic translation constraints~\citep{Zhu2017CycleGAN,Park2020CUT} or remote-sensing-specific biases for geometry, frequency, multiscale representation, misalignment, and perturbation robustness~\citep{Zhang2022PDE,Lee2023CFCASET,Yang2025S3OIL}. Other approaches incorporate auxiliary observations or pretrained knowledge~\citep{Lee2023CFCASET,Yang2025S3OIL}. More recently, general-purpose generative priors such as Stable Diffusion have been transferred to remote sensing translation~\citep{Do2026CDiffSET,Zhao2025RLIDM}, but are typically treated as initialization or backbones. Target-domain generative priors therefore remain underexplored as independently pretrained, evaluated, and scaled components of cross-modal translation.

However, separating these learning roles does not make their independently trained components directly composable. Although unpaired generative pretraining can estimate the target marginal~\citep{Ho2020DDPM,Rombach2022LDM}, marginals alone do not identify cross-modal dependence~\citep{Xie2023Identifiability}. Even a high-quality unconditional prior may therefore lack task-relevant target modes or representations readily controllable by source observations. Moreover, freezing the backbone transfers task-specific correction to the conditioning module, whereas full updating risks perturbing the learned prior. Our conditional-score and denoising-risk analyses make this tension explicit: replacing the task-specific target field $U_p$ with a pretrained field $U_q$ yields the optimal paired-stage correction $C_p+(U_p-U_q)$, comprising both cross-modal conditioning and prior-mismatch compensation~\citep{Chao2022DLSM}. Naive two-stage training may thus shift rather than reduce the burden on paired learning. Effective decoupling requires both adequate target-domain support and source-conditioned controllability without eroding the prior.

To meet these requirements, we propose Learning the Target Priors Before Image Translation (LTP-BIT), a two-stage framework that decouples target-distribution modeling from cross-modal dependence learning. As illustrated in Fig.~\ref{fig:framework}(b) and (c), the first stage learns a target-domain generative prior from large-scale unpaired imagery and examines its task support by varying the pretraining data scale, model capacity, and target-domain composition. The second stage freezes the pretrained backbone weights and uses limited paired data to learn the task-specific adaptation required for cross-modal translation. To make the independently learned prior effectively controllable, we introduce the Prior-Derived Dual-Stream Asymmetric Reference Transformer (P-DART). Through asymmetric interaction between a generation stream and a source-reference stream, P-DART first updates the reference representation according to the current generation state, and then allows the generation stream to retrieve denoising-relevant source information. Only the reference branch and LoRA adapters~\citep{Hu2022LoRA} are optimized, preserving the pretrained prior while concentrating task-specific learning on cross-modal conditional adaptation.

Our contributions are threefold. First, we formulate a prior-first paradigm that decouples target-distribution modeling from cross-modal dependence learning. Our conditional-score and denoising-risk analyses reveal that prior mismatch becomes an additional burden on conditional adaptation, establishing target-domain support and conditional controllability as two requirements for effective decoupling.
Second, we conduct a controlled study of target-prior suitability and scaling across pretraining domains, data scales, model capacities, and data compositions, complemented by unpaired diagnostics of prior support. The results show that prior matching and scaling primarily benefit target-domain realism, while instance-level fidelity relies more strongly on paired conditional adaptation.
Third, we introduce P-DART, a parameter-efficient architecture for cross-modal control of pretrained DiT priors. Using only \(9.81\%\) task-specific parameters, LTP-BIT achieves state-of-the-art performance across SAR-to-RGB and NIR-to-RGB benchmarks and, on QXS-SAROPT, retains near-full-data instance fidelity with only \(25\%\) of the paired samples.

\section{Related Work}

\textbf{Image-to-Image Translation.}
Image translation methods primarily learn task-specific mappings under different supervision regimes. Paired methods employ adversarial, transport, or denoising objectives~\citep{Isola2017Pix2Pix,Li2023BBDM,Liu2023I2SB,Ho2020DDPM}, whereas semi-supervised and unsupervised methods exploit unpaired data through consistency, cycle, latent-space, or contrastive constraints~\citep{Yang2025S3OIL,Zhu2017CycleGAN,Park2020CUT}. Across these regimes, target-distribution modeling remains embedded in translation training rather than being independently pretrained and scaled.

\textbf{Remote Sensing Cross-Modal Image Translation.}
Existing methods mitigate underdetermination by expanding training observations, encoding task-specific inductive biases, or transferring pretrained knowledge. They respectively exploit large-scale multimodal pairs, unpaired source imagery, or auxiliary modalities~\citep{Chen2026Any2Any,Yang2025S3OIL,Lee2023CFCASET}; impose cycle, structural, spatial-frequency, global-local, or misalignment-robust constraints~\citep{Yang2022FGGAN,Zhang2022PDE,Wang2026CDTSDE,Lee2023CFCASET}; or transfer discriminative representations and general-purpose generative priors~\citep{He2025DOGAN,Do2026CDiffSET,Zhao2025RLIDM}. Yet target-distribution modeling either remains coupled with task-specific mapping or is inherited from a general-purpose backbone. Target-domain generative priors remain underexplored as independently pretrained and scalable components, particularly regarding how their pretraining data, capacity, and distributional coverage affect downstream translation.

\textbf{Conditional Adaptation of Pretrained Generative Models.}
Pretrained generative models incorporate conditions through residual injection, key-value attention, or joint attention. Residual methods attach external control branches to the generative backbone~\citep{Zhang2023ControlNet,Cao2026RelaCtrl}, while attention-based methods either expose condition features as additional keys and values or jointly update condition and generation tokens~\citep{Ye2023IPAdapter,Esser2024SD3,Tan2025OminiControl}. Remote sensing systems similarly adapt Stable Diffusion with ControlNet-style branches for imagery and metadata~\citep{Khanna2024DiffusionSat,Tang2024CRSDiff}. However, existing designs do not explicitly separate generation-conditioned reference updating from subsequent source-information retrieval. Generation-state-aware control of independently learned target priors therefore remains underexplored.

\section{Method}

\subsection{Conditional Score Decomposition: Why Learn the Target-Domain Prior Before Translation?}
\label{sec:bayes}

\paragraph{Target-Prior and Cross-Modal Decomposition}

Let $X$ denote a source-modality observation, $Y$ a target-modality image, and $Z=E(Y)$ the target latent variable. 
We define $Z_t=\alpha_t Z+\sigma_t\varepsilon$ as the noisy state at time $t$, where $\varepsilon\sim\mathcal N(0,I)$.

Given a source observation $X=x$, the score of the ideal conditional distribution satisfies
\begin{equation}
\nabla_{z_t}\log p_t^\star(z_t\mid x)
=
\nabla_{z_t}\log p_t^\star(z_t)
+
\nabla_{z_t}\log p_t^\star(x\mid z_t).
\label{eq:score}
\end{equation}
The two terms correspond to the target-domain prior learned from unpaired target images and the source-dependent likelihood learned from paired samples, respectively. This decomposition motivates LTP-BIT to learn the target-domain prior before paired cross-modal adaptation.

To further analyze how the pretrained prior affects the subsequent translation stage, let
$\omega=(Z_t,t)$ and let
$\xi=\psi_t(Z,\varepsilon)$
denote the flow-matching prediction target. Let \(p\) and \(q\) denote the paired-translation and unpaired-pretraining distributions, respectively; they share the same latent space, noising path, and prediction parameterization. Under this Gaussian path, flow-matching prediction fields are time-dependent affine transforms of the corresponding score fields. We define \(F_p^\star=\mathbb E_p[\xi\mid\omega,X]\), \(U_p=\mathbb E_p[\xi\mid\omega]\), \(C_p=F_p^\star-U_p\), and \(U_q=\mathbb E_q[\xi\mid\omega]\) as the optimal paired conditional field, task-specific target-domain prior field, cross-modal conditional component, and pretrained prior field, respectively.

Taking the pretrained prior field $U_q$ as a reference, we denote the overall correction introduced during the paired translation stage by $g$, such that
$F=U_q+g$.
Under the population squared risk, the optimal correction satisfies
\begin{equation}
g^\star=C_p+(U_p-U_q).
\label{eq:fixed-prior}
\end{equation}
Thus, paired translation must introduce the cross-modal information \(C_p\) while compensating for the prior discrepancy \(U_p-U_q\); the latter decreases as \(U_q\) approaches \(U_p\). P-DART and backbone LoRA jointly perform this adaptation. The score-to-prediction-field transformation and squared-risk derivation are provided in Section A.1 of the supplementary material.

\begin{figure}[t]
\centering
\includegraphics[width=\columnwidth]{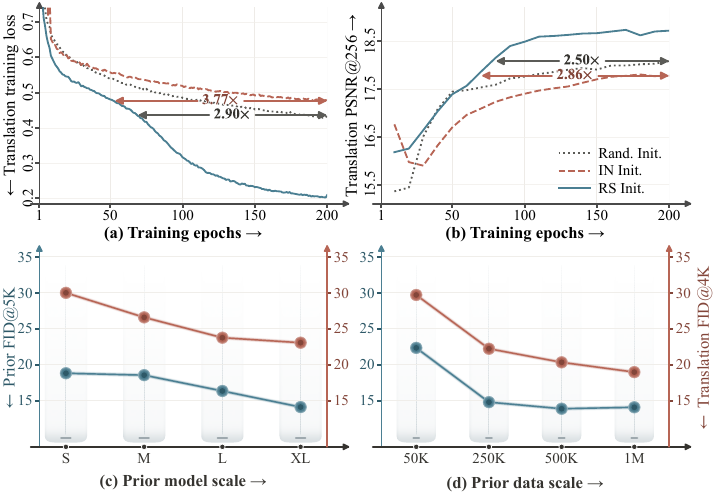}
\caption{Target-domain prior initialization and scaling on QXS-SAROPT. Panels (a, b) compare random, ImageNet, and remote-sensing initialization under full-parameter fine-tuning; arrows report the epoch speedups achieved by remote-sensing initialization in reaching the respective final baseline levels. Under frozen-prior conditional adaptation, panel (c) varies the prior-backbone size with fixed DDT-head capacity, while panel (d) varies a strictly nested pretraining pool with the L prior and adaptation protocol fixed.}
\label{fig:prior-scale}
\end{figure}

\paragraph{Adaptation Efficiency and Scaling Effects of Target-Domain Priors}

To what extent can a target-domain prior compensate for inadequate target-domain distribution learning when paired data are limited? We first compare different initializations under full-parameter fine-tuning. As shown in Figure~\ref{fig:prior-scale}(a, b), remote-sensing prior initialization reaches the final training-loss and translation-PSNR levels of the random and ImageNet baselines in fewer epochs. However, because full-parameter fine-tuning updates the generative backbone, this comparison cannot isolate the contribution of a fixed target-domain prior. We therefore separately scale the pretraining data pool and prior-backbone capacity under a common paired-data budget and frozen-prior adaptation protocol. The prior adopts the \(\mathrm{DiT}^{\mathrm{DH}}\) architecture~\citep{Zheng2025RAE}. In the prior-backbone scaling experiment, all pretrained prior weights are frozen, and conditional adaptation is restricted to the DDT head, where the P-DART reference stream and LoRA modules remain trainable; their counterparts in the DiT backbone are disabled. Because all four prior configurations use identical DDT-head architectures and sizes, the conditioning architecture and number of trainable parameters remain unchanged. The full P-DART design is described in Section~\ref{sec:control}.

Under a fixed conditional-adaptation protocol, Figure~\ref{fig:prior-scale}(c, d) shows closely aligned trends in prior-generation and translation FID as the prior backbone and pretraining data are scaled.

However, this alignment does not extend to pixel-wise fidelity. Under fixed conditioning capacity, Table~\ref{tab:prior-scale-summary} shows that CMMD improves overall with model and data scaling, whereas PSNR varies non-monotonically with model size and decreases with increasing pretraining data. Thus, a better fit to the target distribution does not necessarily yield higher pixel-wise fidelity. Full configurations are provided in Supplementary Section B.2.

\begin{table*}[t]
\centering
\footnotesize
\renewcommand{\arraystretch}{1.08}
\begin{tabular}{@{}c@{\hspace{12pt}}c@{\hspace{12pt}}c@{}}

{\setlength{\tabcolsep}{3.5pt}
\begin{tabular}[t]{@{}lccc@{}}
\toprule
Prior & Params & CMMD$\downarrow$ & PSNR$\uparrow$ \\
\midrule
S  & 10.0M & 0.265 & \cellcolor{black!24}15.569 \\
M  & 10.0M & \cellcolor{black!8}0.254 & \cellcolor{black!8}15.422 \\
L  & 10.0M & \cellcolor{black!24}0.232 & \cellcolor{black!16}15.437 \\
XL & 10.0M & \cellcolor{black!24}0.232 & 15.009 \\
\bottomrule
\end{tabular}}
&
{\setlength{\tabcolsep}{3.5pt}
\begin{tabular}[t]{@{}lccc@{}}
\toprule
Prior & Params & CMMD$\downarrow$ & PSNR$\uparrow$ \\
\midrule
S  & 10.6M & 0.273 & 15.888 \\
M  & 14.6M & \cellcolor{black!8}0.258 & \cellcolor{black!8}15.965 \\
L  & 22.0M & \cellcolor{black!16}0.248 & \cellcolor{black!16}16.109 \\
XL & 45.4M & \cellcolor{black!24}0.226 & \cellcolor{black!24}16.194 \\
\bottomrule
\end{tabular}}
&
{\setlength{\tabcolsep}{3.5pt}
\begin{tabular}[t]{@{}lcc@{}}
\toprule
Data & CMMD$\downarrow$ & PSNR$\uparrow$ \\
\midrule
50K  & 0.253 & \cellcolor{black!24}16.093 \\
250K & \cellcolor{black!8}0.231 & \cellcolor{black!16}16.009 \\
500K & \cellcolor{black!16}0.226 & \cellcolor{black!8}15.912 \\
1M   & \cellcolor{black!24}0.224 & 15.845 \\
\bottomrule
\end{tabular}}
\\[4pt]

{\scriptsize
\shortstack{
\textbf{(a) Prior-Model Scaling}\\
\textbf{(Fixed DDT-Head Conditioning)}
}}
&
{\scriptsize
\shortstack{
\textbf{(b) Prior-Model Scaling}\\
\textbf{(Full P-DART)}
}}
&
{\scriptsize
\shortstack{
\textbf{(c) Pretraining-Data}\\
\textbf{Scaling}
}}

\end{tabular}

\caption{Prior and data scaling. Panels (a) and (b) apply P-DART conditioning to the DDT head only and the full backbone, respectively; trainable parameters are therefore fixed in (a) and scale with the backbone in (b). Panel (c) scales nested pretraining data with the L prior and adaptation protocol fixed. Darker shading denotes better results.}
\label{tab:prior-scale-summary}
\end{table*}

\subsection{Target Prior Coverage: What Kind of Prior Truly Benefits Translation?}\label{sec:coverage}

The preceding results show that improved distribution-level quality does not necessarily yield higher pixel-wise fidelity, suggesting that aggregate distributional distance alone is insufficient to characterize how a target prior benefits translation. We therefore examine local structural compatibility and target-manifold coverage as two complementary properties.

\textbf{Local Feature-Structure Compatibility Hypothesis:} Complex visual patterns in the target domain are jointly formed by local features and the ways in which those features are combined. If the prior and target distributions are highly compatible with respect to these local structures, cross-modal adaptation can realize the conditional mapping by selecting and recombining existing representations, thereby reducing the need to relearn fundamental visual structures in the target domain. We therefore use one-step target-feature recovery as a proxy for local structural compatibility. Starting from the same perturbed target latents, we perform unconditional recovery with the frozen prior and conditional recovery with paired SAR observations, and measure the reduction in spatially aligned DINOv3 patch-feature discrepancy. Prior TFRR and Translation TFRR denote the fractions of this discrepancy removed by unconditional and conditional recovery, respectively.

\begin{figure}[t]
\centering
\includegraphics[width=\columnwidth]{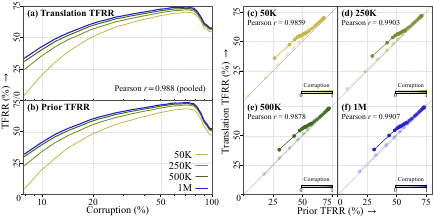}
\caption{Prior and Translation TFRR across nested pretraining-pool sizes. Panels (a) and (b) show Translation and Prior TFRR, respectively, versus normalized target-feature perturbation; panels (c)--(f) show their correspondence for 50K, 250K, 500K, and 1M pretraining samples. Color denotes perturbation magnitude; the dashed line marks equality.}
\label{fig:coverage}
\end{figure}

Figure~\ref{fig:coverage} shows that Prior TFRR and Translation TFRR increase together with pretraining scale and remain strongly correlated across and within scales. The correlation analysis excludes extremely weak perturbations as detailed in Supplementary Section D.1.

\textbf{Target Feature Manifold Coverage Hypothesis:} The target feature manifold reflects the diversity and spatial organization of target-domain patterns. We hypothesize that broader prior coverage makes a wider range of target representations available during conditional adaptation, reducing the need to recover missing modes from limited paired data. Because aggregate distributional metrics do not reveal how generated features occupy this manifold, we additionally measure Coverage and Density in the DINOv2 feature space~\citep{Naeem2020Reliable}, which quantify target-manifold coverage and normalized local overlap with real target features, respectively.

\begin{table}[t]
\centering
\caption{Target-manifold metrics and paired translation performance across nested pretraining-pool sizes.}
\label{tab:manifold-coverage}
\scriptsize
\renewcommand{\arraystretch}{1.08}
\setlength{\tabcolsep}{1.6pt}
\resizebox{\columnwidth}{!}{%
\begin{tabular}{@{}lcccccc@{}}
\toprule
\multirow{2}{*}{Data scale}
& \multicolumn{2}{c}{Manifold}
& \multicolumn{2}{c}{Distribution-level}
& \multicolumn{2}{c}{Instance-level} \\
\cmidrule(lr){2-3}\cmidrule(lr){4-5}\cmidrule(lr){6-7}
& Cov.@3 (\%)$\uparrow$
& Dens.@3$\uparrow$
& FID$\downarrow$
& CMMD$\downarrow$
& PSNR$\uparrow$
& LPIPS$\downarrow$ \\
\specialrule{\lightrulewidth}{\aboverulesep}{0pt}
50K  & 7.675 & \cellcolor{black!24}0.062 & 29.70 & 0.253 & \cellcolor{black!24}16.093 & 0.469 \\
250K & \cellcolor{black!8}7.750 & \cellcolor{black!16}0.055 & \cellcolor{black!8}22.23 & \cellcolor{black!8}0.231 & \cellcolor{black!16}16.009 & \cellcolor{black!16}0.460 \\
500K & \cellcolor{black!16}7.925 & 0.052 & \cellcolor{black!16}20.34 & \cellcolor{black!16}0.226 & \cellcolor{black!8}15.912 & \cellcolor{black!24}0.459 \\
1M   & \cellcolor{black!24}8.000 & \cellcolor{black!8}0.053 & \cellcolor{black!24}18.98 & \cellcolor{black!24}0.224 & 15.845 & \cellcolor{black!8}0.461 \\
\specialrule{\heavyrulewidth}{0pt}{\belowrulesep}
\end{tabular}
}
\end{table}

Across the pretraining-pool expansion, Coverage and distribution-level metrics improve, while Density and instance-level fidelity do not improve consistently (Table~\ref{tab:manifold-coverage}). At a fixed pool size, increasing the proportion of QXS targets improves Coverage, Density, and the translation metrics overall (Table~\ref{tab:target-enrichment}). Together with the TFRR results, these findings indicate that prior utility depends not on scale alone but also on local feature-structure compatibility and task-relevant target-manifold coverage. Motivated by this pattern, our final corpus combines large-scale remote-sensing imagery for domain diversity with target images from each benchmark's training split for task-specific support.

\begin{table}[t]
\centering
\caption{Target-manifold metrics and paired translation performance under fixed-pool target-domain enrichment. All priors are initialized from the same model trained for 300 epochs on a 50K-sample pool and are then trained for 100 additional epochs; Coverage and Density are computed from 5K generated samples.}
\label{tab:target-enrichment}
\scriptsize
\renewcommand{\arraystretch}{1.08}
\setlength{\tabcolsep}{1.6pt}
\resizebox{\columnwidth}{!}{%
\begin{tabular}{@{}lcccccc@{}}
\toprule
\multirow{2}{*}{QXS}
& \multicolumn{2}{c}{Manifold}
& \multicolumn{2}{c}{Distribution-level}
& \multicolumn{2}{c}{Instance-level} \\
\cmidrule(lr){2-3}\cmidrule(lr){4-5}\cmidrule(lr){6-7}
& Cov.@3 (\%)$\uparrow$
& Dens.@3$\uparrow$
& FID$\downarrow$
& CMMD$\downarrow$
& PSNR$\uparrow$
& LPIPS$\downarrow$ \\
\midrule
0
& 7.075
& 0.0481
& 28.18
& \cellcolor{black!8}0.239
& \cellcolor{black!8}16.187
& 0.467 \\
4K
& \cellcolor{black!8}21.275
& \cellcolor{black!8}0.1317
& \cellcolor{black!8}27.75
& 0.241
& 16.138
& \cellcolor{black!8}0.464 \\
8K
& \cellcolor{black!16}29.075
& \cellcolor{black!16}0.2028
& \cellcolor{black!24}26.41
& \cellcolor{black!24}0.234
& \cellcolor{black!16}16.220
& \cellcolor{black!16}0.460 \\
16K
& \cellcolor{black!24}36.900
& \cellcolor{black!24}0.2938
& \cellcolor{black!16}26.75
& \cellcolor{black!16}0.236
& \cellcolor{black!24}16.268
& \cellcolor{black!24}0.457 \\
\bottomrule
\end{tabular}
}
\end{table}

\subsection{Conditional Control}\label{sec:control}

\begin{figure}[t]
\centering
\includegraphics[width=\columnwidth]{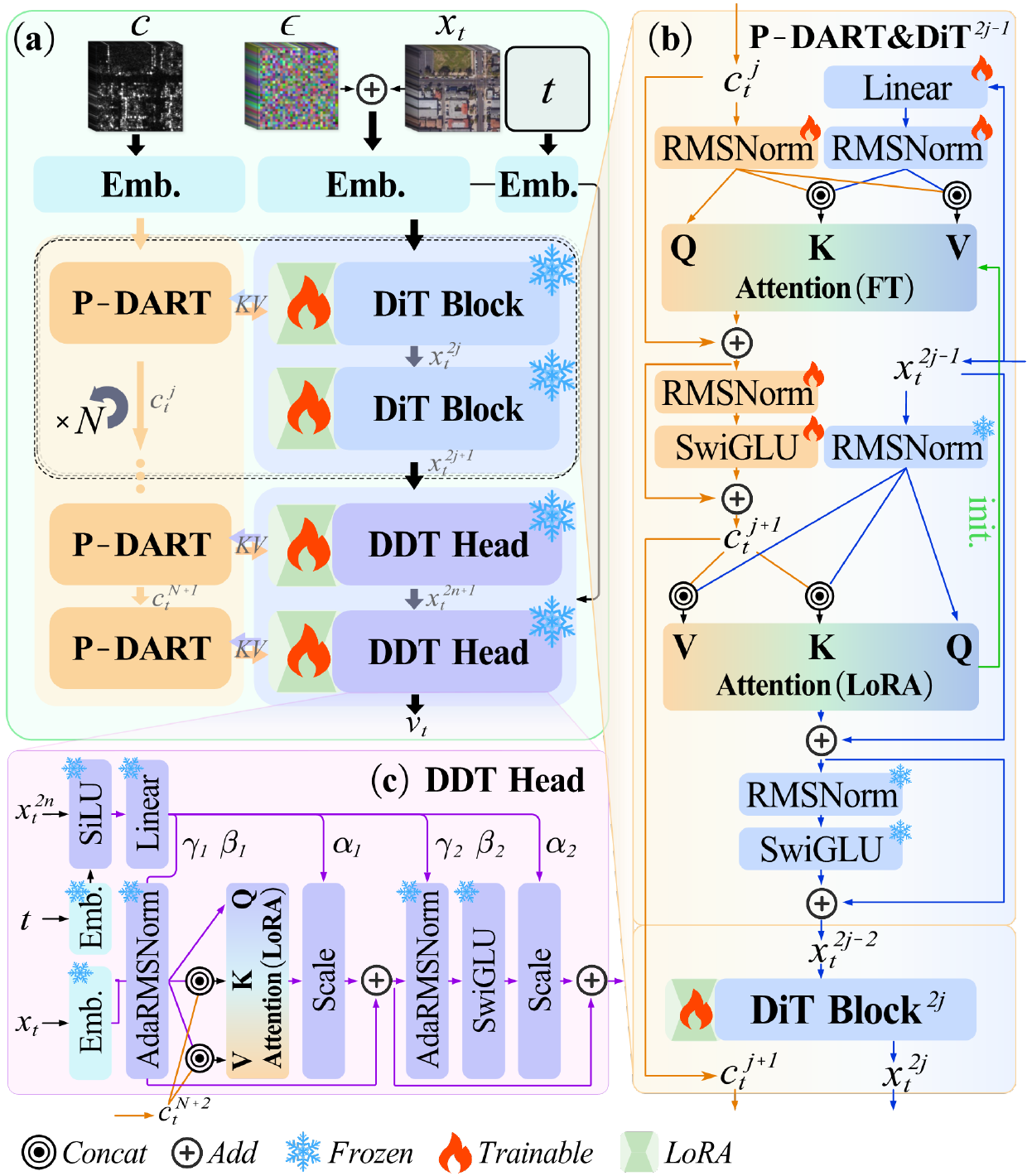}
\caption{Dual-Stream Conditional Adaptation Architecture of P-DART. (a) Propagation path of the source-modality reference stream through the \(\mathrm{DiT}^{\mathrm{DH}}\) architecture, comprising a DiT backbone and a DDT head; (b) bidirectional attention interactions between the reference and generation streams in adjacent DiT blocks, where the reference-stream attention is fully trainable, whereas the generation-stream attention is adapted using LoRA; and (c) architecture of the DDT head and the locations where LoRA is applied.}
\label{fig:model-architecture}
\end{figure}

Paired conditional adaptation must establish source-dependent control over generation and may also need to accommodate residual mismatch between the pretrained prior and the task-specific target domain. Full-parameter fine-tuning exposes the generative backbone to task-specific gradients and may perturb its pretrained generative capabilities~\citep{Zhang2023ControlNet}. Guided by Equations~\eqref{eq:score} and~\eqref{eq:fixed-prior}, we keep the pretrained base weights fixed, introduce LoRA into the generative backbone to provide capacity for task-specific prior adjustment, and employ the fully trainable P-DART reference stream to inject source-dependent conditioning into the generation process.

The \(\mathrm{DiT}^{\mathrm{DH}}\) architecture consists of a DiT backbone and a shallow, wide DDT head. As shown in Figure~\ref{fig:model-architecture}(b), P-DART sequentially performs two attention updates between the generation stream \(x_t^j\) and the source-modality reference stream \(c_t^j\). The reference stream first uses \(c_t^j\) as the query and jointly attends to the source condition and the current generation state:
\begin{equation}
\widetilde c_t^j=
\operatorname{Attn}_{\mathrm{ref}}
\left(
Q_c(c_t^j),K_c([c_t^j;x_t^j]),V_c([c_t^j;x_t^j])
\right).
\label{eq:pdart-reference}
\end{equation}
After \(c_t^{j+1}\) is obtained through the residual connection and feed-forward network, the generation stream uses \(x_t^j\) as the query and the updated reference representation as additional keys and values:
\begin{equation}
\widetilde x_t^j=
\operatorname{Attn}_{g}
\left(
Q(x_t^j),K([x_t^j;c_t^{j+1}]),V([x_t^j;c_t^{j+1}])
\right).
\label{eq:pdart-generation}
\end{equation}
Here, \([\cdot\,;\cdot]\) denotes token concatenation.

P-DART uses backbone-derived initialization to provide an optimization starting point aligned with the generative backbone~\citep{Zhang2023ControlNet}. Specifically, we copy the first \(25\%\) of the attention heads together with the corresponding \(Q/K/V/O\) projection parameters from the matching backbone layers and train all parameters of the reference stream. To enable a limited number of control modules to cover different representational stages of the backbone, we deploy P-DART at intervals throughout the DiT backbone. This interleaved configuration achieves better generation quality and more effective conditional control with the same number of control modules~\citep{Lin2025CtrlAdapter,Yu2025UniCon}. The shallower DDT head instead performs conditional adaptation at every layer.

During conditional adaptation, we apply a noise curriculum by probabilistically replacing the target-side conditioning input to the generative backbone with pure noise to strengthen cross-modal dependence modeling. We use the same flow-matching objective as in pretraining and jointly optimize P-DART and the backbone LoRA modules.

To test whether fixed conditioning capacity limits the instance-level gains from stronger priors, we jointly scale P-DART and the prior backbone from S to XL. Unlike prior scaling under fixed conditioning capacity in Table~\ref{tab:prior-scale-summary}(a), Table~\ref{tab:prior-scale-summary}(b) shows consistently decreasing CMMD and increasing PSNR, indicating that stronger priors require commensurate cross-modal conditioning capacity.

\section{Experiments}

\begin{table*}[t]
\centering
\footnotesize
\setlength{\tabcolsep}{1.5pt}
\caption{Comparison on QXS-SAROPT and SpaceNet6. Pretrained Prior reports the pretraining corpus and whether the prior remains frozen or is further trained during task adaptation. Best results are in \textbf{bold}, and second-best results are \underline{underlined}.}
\label{tab:main}
\makebox[\textwidth][c]{%
\begin{tabular}{@{}llll|ccccc||ccccc@{}}
\specialrule{\heavyrulewidth}{0pt}{0pt}
\rowcolor{gray!12} \multicolumn{1}{@{}>{\columncolor{gray!12}[0pt][\tabcolsep]}l}{} & & & & \multicolumn{5}{c||}{\rule{0pt}{2.3ex}QXS-SAROPT} & \multicolumn{5}{>{\columncolor{gray!12}[\tabcolsep][0pt]}c@{}}{SpaceNet6} \\
\hhline{>{\arrayrulecolor{gray!12}}---->{\arrayrulecolor{black}}----------}
\rowcolor{gray!12} \multicolumn{1}{@{}>{\columncolor{gray!12}[0pt][\tabcolsep]}l}{\multirow{-2}{*}{Method}} & \multirow{-2}{*}{Venue} & \multirow{-2}{*}{Type} & \multicolumn{1}{l|}{\multirow{-2}{*}{\shortstack{Pretrained\\Prior}}} &\rule[-1ex]{0pt}{3.4ex}PSNR$\uparrow$ & SSIM$\uparrow$ & LPIPS$\downarrow$ & FID$\downarrow$ & CMMD$\downarrow$ & PSNR$\uparrow$ & SSIM$\uparrow$ & LPIPS$\downarrow$ & FID$\downarrow$ & \multicolumn{1}{>{\columncolor{gray!12}[\tabcolsep][0pt]}c@{}}{CMMD$\downarrow$} \\
\specialrule{0.05em}{0pt}{0pt}
\specialrule{0.05em}{1.5pt}{2.5pt}
\multicolumn{14}{@{}l}{\textcolor{black!60}{\textit{No pretrained generative prior}}} \\[1.5pt]
CycleGAN & ICCV\textsubscript{17}    & \multirow{3}{*}{G} & --       & 13.012 & 0.258 & 0.618 & 170.70 & 2.521 & 15.052 & 0.253 & 0.572 & 216.20 & 2.007 \\
CFCA-SET & TGRS\textsubscript{23}    &  & --              & \underline{15.660} & \best{0.358} & 0.533 & 183.26 & 2.717 & 16.459 & 0.286 & 0.516 & 239.88 & 4.160 \\
StegoGAN & CVPR\textsubscript{24}    &  & --              & 12.217 & 0.228 & 0.582 & 95.13 & 1.093 & 13.512 & 0.205 & 0.488 & 136.96 & 1.451 \\
\arrayrulecolor{black!25}\cmidrule[0.03em](l{0.07em}r{0.07em}){1-14}\arrayrulecolor{black}
BBDM & CVPR\textsubscript{23}    & B & --               & 14.282 & 0.304 & 0.520 & 103.03  & 1.007 & 16.935 & 0.327 & \underline{0.391} & 60.74 & 0.719 \\ 
\midrule
\multicolumn{14}{@{}l}{\textcolor{black!60}{\textit{Generic-domain prior adaptation}}} \\[1.5pt]
ControlNet & ICCV\textsubscript{23}    & \multirow{3}{*}{D} & LAION\frozenprior & 14.312 & 0.306 & 0.526 & 34.56 & 0.359 & 16.528 & 0.314 & 0.428 & 55.44 & 0.771 \\
Uni-ControlNet & NeurIPS\textsubscript{23} &  & LAION\frozenprior        & 13.924 & 0.295 & 0.530 & 36.20  & 0.374 & 15.959 & 0.287 & 0.451 & 57.49 & 0.781 \\
C-DiffSET & TCSVT\textsubscript{26}   &  & LAION\updatedprior             & 14.256 & 0.310 & 0.519 & 30.58 & 0.350 & 16.591 & 0.320 & 0.437 & 44.16 & 0.633 \\
\midrule
\multicolumn{14}{@{}l}{\textcolor{black!60}{\textit{Target-domain prior pretraining}}} \\[1.5pt]
DiffusionSat$^\dagger$   & ICLR\textsubscript{24} & \multirow{3}{*}{D} & RS-Mix\frozenprior & 14.311 & 0.322 & 0.576 & 31.25 & 0.327 & 15.848 & 0.301 & 0.498 & 54.30 & \underline{0.607} \\
Text2Earth$^\dagger$     & GRSM\textsubscript{25} &  & Git-10M\frozenprior        & 14.737 & 0.332 & \underline{0.508} & \underline{23.15} & \underline{0.233} & \underline{17.576} & \underline{0.345} & 0.397 & \best{35.47} & \best{0.420} \\
Ours-XL        & -- &  & RS-1M\frozenprior                                      & \best{15.956} & \underline{0.351} & \best{0.446} & \best{16.54} & \best{0.201} & \best{18.924} & \best{0.353} & \best{0.293} & \underline{43.76} & 0.808 \\
\bottomrule
\end{tabular}%
}
\par\vspace{1.5pt}
\noindent\parbox{\textwidth}{\scriptsize \textit{Note:} G, B, and D denote GAN, diffusion bridge, and conditional diffusion, respectively. \frozenmark/\updatedmark\ denote whether the generative backbone is frozen or trained during task adaptation. $^\dagger$ denotes adaptations that use only the corresponding released pretrained weights and follow the ControlNet paradigm for task-specific training.}

\end{table*}

\begin{table}[t]
\centering
\footnotesize
\setlength{\tabcolsep}{2.0pt}
\caption{Comparison on Chesapeake. Method variants and notation follow Table~\ref{tab:main}.}
\label{tab:chesapeake}
\begin{tabular}{@{}l|ccccc@{}}
\specialrule{\heavyrulewidth}{0pt}{0pt}
\rowcolor{gray!12}
\multicolumn{1}{@{}>{\columncolor{gray!12}[0pt][\tabcolsep]}l|}{} &
\multicolumn{5}{>{\columncolor{gray!12}[\tabcolsep][0pt]}c@{}}{\rule[-0.8ex]{0pt}{3.0ex}Chesapeake} \\
\hhline{>{\arrayrulecolor{gray!12}}~>{\arrayrulecolor{black}}-----}
\rowcolor{gray!12}
\multicolumn{1}{@{}>{\columncolor{gray!12}[0pt][\tabcolsep]}l|}{\multirow{-2}{*}{Method}} &
\rule[-1.1ex]{0pt}{3.8ex}PSNR$\uparrow$ & SSIM$\uparrow$ & LPIPS$\downarrow$ & FID$\downarrow$ &
\multicolumn{1}{>{\columncolor{gray!12}[\tabcolsep][0pt]}c@{}}{CMMD$\downarrow$} \\
\specialrule{0.05em}{0pt}{2.5pt}
\multicolumn{6}{@{}l}{\textcolor{black!60}{\textit{No pretrained generative prior}}} \\[1.5pt]
CycleGAN       & 15.621 & 0.536 & 0.405 & 83.94 & 1.514 \\
CFCA-SET       & 18.569 & \best{0.572} & 0.351 & 85.82 & 2.388 \\
StegoGAN       & 16.099 & \underline{0.559} & 0.390 & 48.91 & \underline{0.625} \\
BBDM           & 17.801 & 0.437 & 0.355 & 53.29 & 0.828 \\
\midrule
\multicolumn{6}{@{}l}{\textcolor{black!60}{\textit{Generic-domain prior adaptation}}} \\[1.5pt]
ControlNet     & 18.255 & 0.469 & 0.328 & 31.34 & 0.671 \\
Uni-ControlNet & 17.909 & 0.431 & 0.348 & 30.73 & 0.703 \\
C-DiffSET      & \underline{18.730} & 0.474 & 0.320 & 37.17 & 0.710 \\
\midrule
\multicolumn{6}{@{}l}{\textcolor{black!60}{\textit{Target-domain prior pretraining}}} \\[1.5pt]
DiffusionSat$^\dagger$ & 17.895 & 0.458 & 0.397 & 39.53 & 0.635 \\
Text2Earth$^\dagger$   & \best{19.338} & 0.497 & \underline{0.303} & \underline{25.59} & \best{0.483} \\
Ours-XL                & 17.717 & 0.350 & \best{0.290} & \best{18.19} & 0.825 \\
\bottomrule
\end{tabular}
\end{table}

\begin{figure*}[t]
\centering
\includegraphics[width=\textwidth]{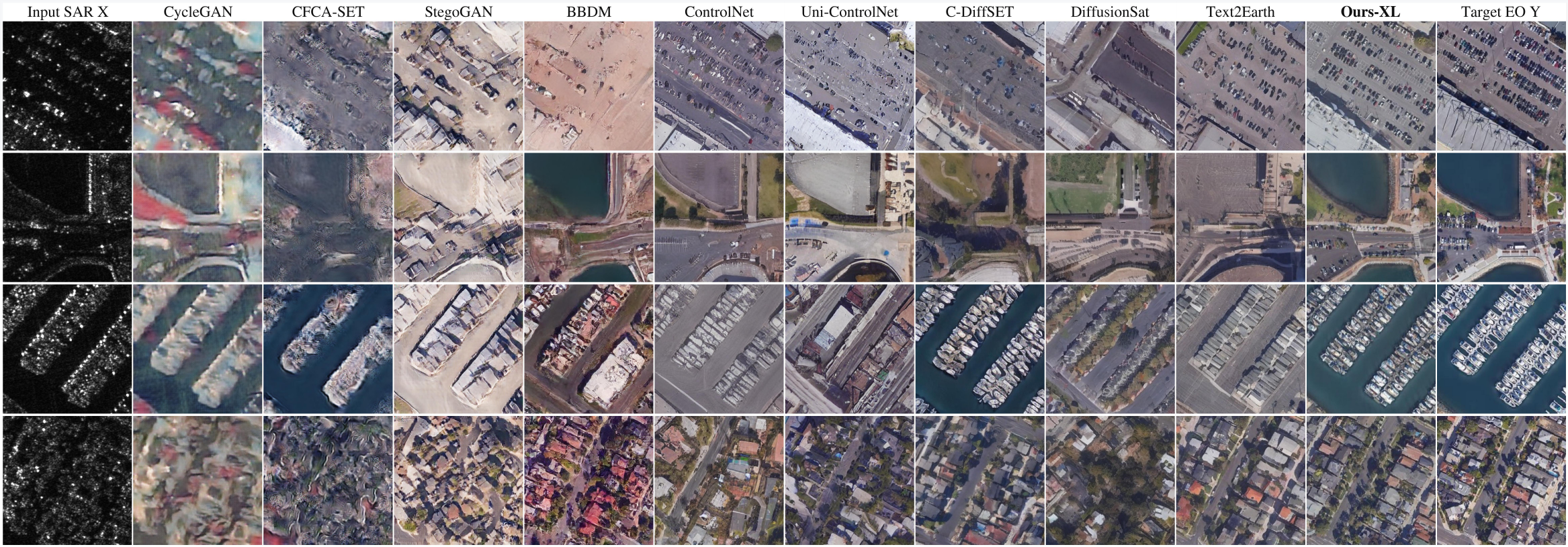}
\caption{Qualitative comparisons on the QXS-SAROPT dataset.}
\label{fig:result1}
\end{figure*}

\textbf{Datasets.}
We evaluate single-polarization SAR-to-RGB, multi-polarization SAR-to-RGB, and NIR-to-RGB translation on QXS-SAROPT, SpaceNet6, and Chesapeake, respectively. QXS-SAROPT contains 20,000 registered $256\times256$ Gaofen-3 SAR and Google Earth RGB pairs at $1\,\mathrm{m}$ resolution, split into 16,000 training and 4,000 test pairs. SpaceNet6 contains 3,401 pairs of $900\times900$ SAR and optical tiles. We partition these tiles into training and test sets at a fixed ratio of $4{:}1$. Overlapping $4\times4$ cropping and black-border removal then yield 20,168 training and 5,048 test pairs. For Chesapeake, we apply non-overlapping cropping, cross-state deduplication, and semantic, random, and information-density sampling to six-state four-band NAIP imagery. This produces training and test sets at a ratio of $4{:}1$, comprising 16,000 and 4,000 $256\times256$ NIR-to-RGB pairs, respectively. Further construction details are provided in Supplementary Section C.2.

\textbf{Target-Domain Prior Pretraining Data.}
For the final model, the shared pretraining corpus combines one million unpaired RGB images selected from Git-10M~\cite{Liu2025Text2Earth} with target images from the training splits of the three benchmarks. Benchmark targets are used without source counterparts or pairing information, and all test images are excluded. The Git-10M filtering procedure is detailed in Supplementary Section C.1.

\textbf{Evaluation Metrics.}
We assess content fidelity using PSNR~\cite{Hore2010ImageQuality}, SSIM~\cite{Wang2004ImageQuality}, and LPIPS~\cite{Zhang2018Perceptual}, and target-domain realism using FID~\cite{Heusel2017TTUR}. We additionally report CMMD, which provides more stable estimates on small evaluation sets~\cite{Jayasumana2024RethinkingFID}.

\textbf{Implementation Details.}
All experiments use four NVIDIA RTX A6000 GPUs. The final $\mathrm{DiT}^{\mathrm{DH}}$-XL prior is pretrained on the above corpus for 100 epochs. During downstream translation, the pretrained base weights remain fixed, while the P-DART conditioning branch and backbone LoRA parameters are optimized for 80 epochs using AdamW with a global batch size of 256. Inference uses 50 sampling steps.

\textbf{Comparison Methods.}
We compare nine methods grouped by pretraining source. Methods without pretrained generative priors are CycleGAN~\cite{Zhu2017CycleGAN}, CFCA-SET~\cite{Lee2023CFCASET}, StegoGAN~\cite{Wu2024StegoGAN}, and BBDM~\cite{Li2023BBDM}. Generic-domain prior methods are ControlNet~\cite{Zhang2023ControlNet}, Uni-ControlNet~\cite{Zhao2023UniControlNet}, and C-DiffSET~\cite{Do2026CDiffSET}. We further adapt the remote-sensing optical priors of DiffusionSat~\cite{Khanna2024DiffusionSat} and Text2Earth~\cite{Liu2025Text2Earth} using the same ControlNet paradigm.

\subsection{Comparison with Existing Methods}

Tables~\ref{tab:main} and~\ref{tab:chesapeake} show that our method achieves strong overall performance across all three datasets, with consistent advantages in perceptual similarity and distributional realism. Methods pretrained on target-domain imagery generally outperform those adapted from generic-domain priors, demonstrating the importance of alignment between the prior and target domains.

\begin{figure}[t]
\centering
\includegraphics[width=\linewidth]{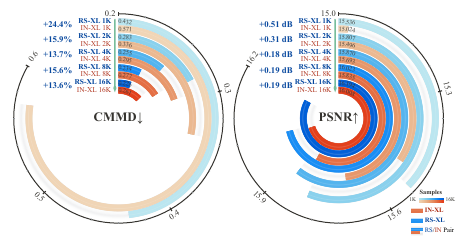}
\caption{Effect of paired-sample scale under matched trainable parameters and 5,040 training steps. Concentric rings report CMMD (left; lower is better) and PSNR (right; higher is better) for RS-XL and IN-XL. From outer to inner, ring pairs correspond to 1K, 2K, 4K, 8K, and 16K paired samples. Annotations report the relative CMMD reduction and absolute PSNR gain of RS-XL over IN-XL.}
\label{fig:budget}
\end{figure}

\subsection{Ablation Studies}\label{sec:ablation}

\textbf{Effect of Paired-Sample Scale.}
We compare RS-XL and IN-XL using 1K, 2K, 4K, 8K, and 16K paired samples under matched trainable parameters and 5,040 training steps. For both priors, increasing paired data consistently raises PSNR and reduces CMMD (Figure~\ref{fig:budget}). RS-XL outperforms IN-XL at every budget, with the largest margins at 1K, showing that target-domain pretraining is particularly beneficial when paired supervision is scarce.

\textbf{Effect of Fine-Tuning Strategy.}
Table~\ref{tab:prior-adaptation} compares P-DART with full fine-tuning and parameter-matched PEFT baselines. Using about one-ninth as many trainable parameters, P-DART reduces FID and CMMD and improves LPIPS relative to full fine-tuning, while retaining close PSNR and SSIM. Against parameter-matched LoRA~\citep{Hu2022LoRA} and DoRA~\citep{Liu2024DoRA}, it improves PSNR, SSIM, and LPIPS, with similar FID and CMMD.

\begin{table}[t]
\centering
\footnotesize
\renewcommand{\arraystretch}{1.08}
\setlength{\tabcolsep}{2pt}
\caption{Full fine-tuning (FT) versus frozen-base adaptation. LoRA and DoRA use the Addition pathway.}
\label{tab:prior-adaptation}
\resizebox{\columnwidth}{!}{%
\begin{tabular}{@{}lccccccc@{}}
\specialrule{\heavyrulewidth}{0pt}{0pt}
\rowcolor{gray!12}
\rule[-1ex]{0pt}{3.6ex}Method & Base & Param & PSNR$\uparrow$ & SSIM$\uparrow$ &
LPIPS$\downarrow$ & FID$\downarrow$ & CMMD$\downarrow$ \\
\specialrule{\lightrulewidth}{0pt}{0pt}
Add.    & FT     & 873M & 16.349 & 0.380 & 0.459 & 55.03 & 0.358 \\
Concat. & FT     & 877M & 16.240 & 0.377 & 0.472 & 62.09 & 0.368 \\
\midrule
LoRA    & Frozen & 95M  & 15.936 & 0.361 & 0.455 & 20.55 & 0.222 \\
DoRA    & Frozen & 95M  & 15.903 & 0.359 & 0.455 & 20.67 & 0.220 \\
\midrule
\textbf{Ours-XL} & Frozen & 95M & 16.194 & 0.362 & 0.445 & 20.68 & 0.226 \\
\bottomrule
\end{tabular}%
}
\end{table}

\section{Conclusion}

We introduced LTP-BIT, a prior-first paradigm that decouples target-domain prior learning from cross-modal dependence learning. Our analysis shows that prior mismatch creates an additional burden for conditional adaptation, while controlled experiments reveal that prior scaling primarily improves target-domain realism and requires sufficient conditioning capacity to benefit instance fidelity. P-DART enables parameter-efficient, generation-aware control of the pretrained DiT prior. LTP-BIT achieves state-of-the-art performance across SAR-to-RGB and NIR-to-RGB benchmarks with only \(9.81\%\) task-specific parameters, and retains near-full-data fidelity on QXS-SAROPT using \(25\%\) of the paired samples. These results establish target generative priors as independently scalable components, allowing scarce paired supervision to focus on cross-modal correspondence.

\bibliography{references}

\begin{thebibliography}{42}
\providecommand{\natexlab}[1]{#1}

\bibitem[{Cao et~al.(2026)Cao, Wang, Ma, Feng, He, Ling, Liu, Lu, Feng, Wang,
  Pei, Shao, Zhang, and Zhang}]{Cao2026RelaCtrl}
Cao, K.; Wang, J.; Ma, A.; Feng, J.; He, X.; Ling, R.; Liu, H.; Lu, J.; Feng,
  W.; Wang, H.; Pei, H.; Shao, Y.; Zhang, Z.; and Zhang, J. 2026.
\newblock {RelaCtrl}: Relevance-Guided Efficient Control for Diffusion
  Transformers.
\newblock In \emph{Proceedings of the AAAI Conference on Artificial
  Intelligence (AAAI)}, volume~40, 2598--2606.

\bibitem[{Chao et~al.(2022)Chao, Sun, Cheng, Lo, Chang, Liu, Chang, Chen, and
  Lee}]{Chao2022DLSM}
Chao, C.-H.; Sun, W.-F.; Cheng, B.-W.; Lo, Y.-C.; Chang, C.-C.; Liu, Y.-L.;
  Chang, Y.-L.; Chen, C.-P.; and Lee, C.-Y. 2022.
\newblock Denoising Likelihood Score Matching for Conditional Score-based Data
  Generation.
\newblock In \emph{Proceedings of the International Conference on Learning
  Representations (ICLR)}.

\bibitem[{Chen et~al.(2026)Chen, Zhang, Wang, Wang, Huang, Li, Guo, Wang, Wang,
  and Du}]{Chen2026Any2Any}
Chen, H.; Zhang, J.; Wang, H.; Wang, S.; Huang, P.; Li, J.; Guo, H.; Wang, D.;
  Wang, Z.; and Du, B. 2026.
\newblock {Any2Any}: Unified Arbitrary Modality Translation for Remote Sensing.
\newblock In \emph{Proceedings of the International Conference on Machine
  Learning (ICML)}.

\bibitem[{Do et~al.(2026)Do, Lee, Lee, and Kim}]{Do2026CDiffSET}
Do, J.; Lee, J.; Lee, S.; and Kim, M. 2026.
\newblock C-DiffSET: Leveraging Latent Diffusion for SAR-to-EO Image
  Translation with Confidence-Guided Reliable Object Generation.
\newblock \emph{IEEE Transactions on Circuits and Systems for Video
  Technology}.
\newblock Forthcoming.

\bibitem[{Esser et~al.(2024)Esser, Kulal, Blattmann, Entezari, M{\"u}ller,
  Saini, Levi, Lorenz, Sauer, Boesel, Podell, Dockhorn, English, and
  Rombach}]{Esser2024SD3}
Esser, P.; Kulal, S.; Blattmann, A.; Entezari, R.; M{\"u}ller, J.; Saini, H.;
  Levi, Y.; Lorenz, D.; Sauer, A.; Boesel, F.; Podell, D.; Dockhorn, T.;
  English, Z.; and Rombach, R. 2024.
\newblock Scaling Rectified Flow Transformers for High-Resolution Image
  Synthesis.
\newblock In \emph{Proceedings of the International Conference on Machine
  Learning (ICML)}, volume 235 of \emph{Proceedings of Machine Learning
  Research}, 12606--12633.

\bibitem[{He et~al.(2025)He, Chen, Shi, Chen, Yang, and Li}]{He2025DOGAN}
He, J.; Chen, L.; Shi, H.; Chen, Y.; Yang, J.; and Li, W. 2025.
\newblock DOGAN: DINO-Based Optical-Prior-Driven GAN for SAR-to-Optical Image
  Translation.
\newblock \emph{IEEE Transactions on Geoscience and Remote Sensing}, 63:
  5220116.

\bibitem[{Heusel et~al.(2017)Heusel, Ramsauer, Unterthiner, Nessler, and
  Hochreiter}]{Heusel2017TTUR}
Heusel, M.; Ramsauer, H.; Unterthiner, T.; Nessler, B.; and Hochreiter, S.
  2017.
\newblock GANs Trained by a Two Time-Scale Update Rule Converge to a Local Nash
  Equilibrium.
\newblock In \emph{Advances in Neural Information Processing Systems
  (NeurIPS)}, volume~30, 6626--6637.

\bibitem[{Ho, Jain, and Abbeel(2020)}]{Ho2020DDPM}
Ho, J.; Jain, A.; and Abbeel, P. 2020.
\newblock Denoising Diffusion Probabilistic Models.
\newblock In \emph{Advances in Neural Information Processing Systems
  (NeurIPS)}, 6840--6851.

\bibitem[{Hor\'e and Ziou(2010)}]{Hore2010ImageQuality}
Hor\'e, A.; and Ziou, D. 2010.
\newblock Image Quality Metrics: PSNR vs. SSIM.
\newblock In \emph{Proceedings of the International Conference on Pattern
  Recognition (ICPR)}, 2366--2369.

\bibitem[{Hu et~al.(2022)Hu, Shen, Wallis, Allen-Zhu, Li, Wang, Wang, and
  Chen}]{Hu2022LoRA}
Hu, E.~J.; Shen, Y.; Wallis, P.; Allen-Zhu, Z.; Li, Y.; Wang, S.; Wang, L.; and
  Chen, W. 2022.
\newblock {LoRA}: Low-Rank Adaptation of Large Language Models.
\newblock In \emph{International Conference on Learning Representations}.

\bibitem[{Huang et~al.(2026)Huang, Zhang, Tang, Xu, Datcu, and
  Han}]{Huang2026GenAI}
Huang, Z.; Zhang, X.; Tang, Z.; Xu, F.; Datcu, M.; and Han, J. 2026.
\newblock Generative Artificial Intelligence Meets Synthetic Aperture Radar: A
  Survey.
\newblock \emph{IEEE Geoscience and Remote Sensing Magazine}, 14(1): 6--48.

\bibitem[{Isola et~al.(2017)Isola, Zhu, Zhou, and Efros}]{Isola2017Pix2Pix}
Isola, P.; Zhu, J.-Y.; Zhou, T.; and Efros, A.~A. 2017.
\newblock Image-to-Image Translation with Conditional Adversarial Networks.
\newblock In \emph{Proceedings of the IEEE Conference on Computer Vision and
  Pattern Recognition (CVPR)}, 1125--1134.

\bibitem[{Jayasumana et~al.(2024)Jayasumana, Ramalingam, Veit, Glasner,
  Chakrabarti, and Kumar}]{Jayasumana2024RethinkingFID}
Jayasumana, S.; Ramalingam, S.; Veit, A.; Glasner, D.; Chakrabarti, A.; and
  Kumar, S. 2024.
\newblock Rethinking FID: Towards a Better Evaluation Metric for Image
  Generation.
\newblock In \emph{Proceedings of the IEEE/CVF Conference on Computer Vision
  and Pattern Recognition (CVPR)}, 9307--9315.

\bibitem[{Khanna et~al.(2024)Khanna, Liu, Zhou, Meng, Rombach, Burke, Lobell,
  and Ermon}]{Khanna2024DiffusionSat}
Khanna, S.; Liu, P.; Zhou, L.; Meng, C.; Rombach, R.; Burke, M.; Lobell, D.~B.;
  and Ermon, S. 2024.
\newblock {DiffusionSat}: A Generative Foundation Model for Satellite Imagery.
\newblock In \emph{Proceedings of the International Conference on Learning
  Representations (ICLR)}.

\bibitem[{Lee et~al.(2023)Lee, Cho, Seo, Kim, Jeong, and Kim}]{Lee2023CFCASET}
Lee, J.; Cho, H.; Seo, D.; Kim, H.-H.; Jeong, J.; and Kim, M. 2023.
\newblock CFCA-SET: Coarse-to-Fine Context-Aware SAR-to-EO Translation With
  Auxiliary Learning of SAR-to-NIR Translation.
\newblock \emph{IEEE Transactions on Geoscience and Remote Sensing}, 61: 1--18.

\bibitem[{Li et~al.(2023)Li, Xue, Liu, and Lai}]{Li2023BBDM}
Li, B.; Xue, K.; Liu, B.; and Lai, Y.-K. 2023.
\newblock BBDM: Image-to-Image Translation with Brownian Bridge Diffusion
  Models.
\newblock In \emph{Proceedings of the IEEE/CVF Conference on Computer Vision
  and Pattern Recognition (CVPR)}, 1952--1961.

\bibitem[{Lin et~al.(2025)Lin, Cho, Zala, and Bansal}]{Lin2025CtrlAdapter}
Lin, H.; Cho, J.; Zala, A.; and Bansal, M. 2025.
\newblock {Ctrl-Adapter}: An Efficient and Versatile Framework for Adapting
  Diverse Controls to Any Diffusion Model.
\newblock In \emph{Proceedings of the International Conference on Learning
  Representations (ICLR)}.

\bibitem[{Liu et~al.(2025)Liu, Chen, Zhao, Zou, and Shi}]{Liu2025Text2Earth}
Liu, C.; Chen, K.; Zhao, R.; Zou, Z.; and Shi, Z. 2025.
\newblock {Text2Earth}: Unlocking Text-Driven Remote Sensing Image Generation
  with a Global-Scale Dataset and a Foundation Model.
\newblock \emph{IEEE Geoscience and Remote Sensing Magazine}, 13(3): 238--259.

\bibitem[{Liu et~al.(2023)Liu, Vahdat, Huang, Theodorou, Nie, and
  Anandkumar}]{Liu2023I2SB}
Liu, G.-H.; Vahdat, A.; Huang, D.-A.; Theodorou, E.~A.; Nie, W.; and
  Anandkumar, A. 2023.
\newblock I2SB: Image-to-Image Schr{\"o}dinger Bridge.
\newblock In \emph{Proceedings of the 40th International Conference on Machine
  Learning (ICML)}, 22042--22062.

\bibitem[{Liu et~al.(2024)Liu, Wang, Yin, Molchanov, Wang, Cheng, and
  Chen}]{Liu2024DoRA}
Liu, S.-Y.; Wang, C.-Y.; Yin, H.; Molchanov, P.; Wang, Y.-C.~F.; Cheng, K.-T.;
  and Chen, M.-H. 2024.
\newblock {DoRA}: Weight-Decomposed Low-Rank Adaptation.
\newblock In \emph{Proceedings of the 41st International Conference on Machine
  Learning}, volume 235 of \emph{Proceedings of Machine Learning Research},
  32100--32121. PMLR.

\bibitem[{Naeem et~al.(2020)Naeem, Oh, Uh, Choi, and Yoo}]{Naeem2020Reliable}
Naeem, M.~F.; Oh, S.~J.; Uh, Y.; Choi, Y.; and Yoo, J. 2020.
\newblock Reliable Fidelity and Diversity Metrics for Generative Models.
\newblock In \emph{Proceedings of the 37th International Conference on Machine
  Learning}, volume 119 of \emph{Proceedings of Machine Learning Research},
  7176--7185.

\bibitem[{Park et~al.(2020)Park, Efros, Zhang, and Zhu}]{Park2020CUT}
Park, T.; Efros, A.~A.; Zhang, R.; and Zhu, J.-Y. 2020.
\newblock Contrastive Learning for Unpaired Image-to-Image Translation.
\newblock In \emph{Proceedings of the European Conference on Computer Vision
  (ECCV)}, 319--345.

\bibitem[{Rombach et~al.(2022)Rombach, Blattmann, Lorenz, Esser, and
  Ommer}]{Rombach2022LDM}
Rombach, R.; Blattmann, A.; Lorenz, D.; Esser, P.; and Ommer, B. 2022.
\newblock High-Resolution Image Synthesis with Latent Diffusion Models.
\newblock In \emph{Proceedings of the IEEE/CVF Conference on Computer Vision
  and Pattern Recognition (CVPR)}, 10674--10685.

\bibitem[{Schmitt and Zhu(2016)}]{Schmitt2016Fusion}
Schmitt, M.; and Zhu, X.~X. 2016.
\newblock Data Fusion and Remote Sensing: An Ever-Growing Relationship.
\newblock \emph{IEEE Geoscience and Remote Sensing Magazine}, 4(4): 6--23.

\bibitem[{Tan et~al.(2025)Tan, Liu, Yang, Xue, and Wang}]{Tan2025OminiControl}
Tan, Z.; Liu, S.; Yang, X.; Xue, Q.; and Wang, X. 2025.
\newblock {OminiControl}: Minimal and Universal Control for Diffusion
  Transformer.
\newblock In \emph{Proceedings of the IEEE/CVF International Conference on
  Computer Vision (ICCV)}, 14940--14950.

\bibitem[{Tang et~al.(2024)Tang, Cao, Hou, Jiang, Liu, and
  Meng}]{Tang2024CRSDiff}
Tang, D.; Cao, X.; Hou, X.; Jiang, Z.; Liu, J.; and Meng, D. 2024.
\newblock {CRS-Diff}: Controllable Remote Sensing Image Generation with
  Diffusion Model.
\newblock \emph{IEEE Transactions on Geoscience and Remote Sensing}, 62: 1--14.

\bibitem[{Wang et~al.(2004)Wang, Bovik, Sheikh, and
  Simoncelli}]{Wang2004ImageQuality}
Wang, Z.; Bovik, A.~C.; Sheikh, H.~R.; and Simoncelli, E.~P. 2004.
\newblock Image Quality Assessment: From Error Visibility to Structural
  Similarity.
\newblock \emph{IEEE Transactions on Image Processing}, 13(4): 600--612.

\bibitem[{Wang, Chen, and Ren(2026)}]{Wang2026CDTSDE}
Wang, Z.; Chen, Y.; and Ren, S. 2026.
\newblock Adaptive Domain Shift in Diffusion Models for Cross-Modality Image
  Translation.
\newblock In \emph{Proceedings of the International Conference on Learning
  Representations (ICLR)}.

\bibitem[{Wu et~al.(2024)Wu, Chen, Mermet, Hurni, Schindler, Gonthier, and
  Landrieu}]{Wu2024StegoGAN}
Wu, S.; Chen, Y.; Mermet, S.; Hurni, L.; Schindler, K.; Gonthier, N.; and
  Landrieu, L. 2024.
\newblock StegoGAN: Leveraging Steganography for Non-Bijective Image-to-Image
  Translation.
\newblock In \emph{Proceedings of the IEEE/CVF Conference on Computer Vision
  and Pattern Recognition (CVPR)}, 7922--7931.

\bibitem[{Xie et~al.(2023)Xie, Kong, Gong, and Zhang}]{Xie2023Identifiability}
Xie, S.; Kong, L.; Gong, M.; and Zhang, K. 2023.
\newblock Multi-Domain Image Generation and Translation with Identifiability
  Guarantees.
\newblock In \emph{Proceedings of the International Conference on Learning
  Representations (ICLR)}.

\bibitem[{Yang et~al.(2025)Yang, Shi, Li, Qiao, Gao, and Wang}]{Yang2025S3OIL}
Yang, X.; Shi, H.; Li, Z.; Qiao, M.; Gao, F.; and Wang, N. 2025.
\newblock S$^3$OIL: Semi-Supervised SAR-to-Optical Image Translation via
  Multi-Scale and Cross-Set Matching.
\newblock \emph{IEEE Transactions on Image Processing}, 34: 6641--6654.

\bibitem[{Yang et~al.(2022)Yang, Wang, Zhao, and Yang}]{Yang2022FGGAN}
Yang, X.; Wang, Z.; Zhao, J.; and Yang, D. 2022.
\newblock FG-GAN: A Fine-Grained Generative Adversarial Network for
  Unsupervised SAR-to-Optical Image Translation.
\newblock \emph{IEEE Transactions on Geoscience and Remote Sensing}, 60:
  5621211.

\bibitem[{Ye et~al.(2023)Ye, Zhang, Liu, Han, and Yang}]{Ye2023IPAdapter}
Ye, H.; Zhang, J.; Liu, S.; Han, X.; and Yang, W. 2023.
\newblock {IP-Adapter}: Text Compatible Image Prompt Adapter for Text-to-Image
  Diffusion Models.
\newblock arXiv:2308.06721.

\bibitem[{Yu et~al.(2025)Yu, Gu, Hu, Li, and Dong}]{Yu2025UniCon}
Yu, F.; Gu, J.; Hu, J.; Li, Z.; and Dong, C. 2025.
\newblock {UniCon}: Unidirectional Information Flow for Effective Control of
  Large-Scale Diffusion Models.
\newblock In \emph{Proceedings of the International Conference on Learning
  Representations (ICLR)}.

\bibitem[{Zhang, Rao, and Agrawala(2023)}]{Zhang2023ControlNet}
Zhang, L.; Rao, A.; and Agrawala, M. 2023.
\newblock Adding Conditional Control to Text-to-Image Diffusion Models.
\newblock In \emph{Proceedings of the IEEE/CVF International Conference on
  Computer Vision (ICCV)}, 3836--3847.

\bibitem[{Zhang et~al.(2022)Zhang, He, Zhang, Yang, Peng, and
  Guo}]{Zhang2022PDE}
Zhang, M.; He, C.; Zhang, J.; Yang, Y.; Peng, X.; and Guo, J. 2022.
\newblock SAR-to-Optical Image Translation via Neural Partial Differential
  Equations.
\newblock In \emph{Proceedings of the Thirty-First International Joint
  Conference on Artificial Intelligence (IJCAI)}, 1644--1650.

\bibitem[{Zhang et~al.(2018)Zhang, Isola, Efros, Shechtman, and
  Wang}]{Zhang2018Perceptual}
Zhang, R.; Isola, P.; Efros, A.~A.; Shechtman, E.; and Wang, O. 2018.
\newblock The Unreasonable Effectiveness of Deep Features as a Perceptual
  Metric.
\newblock In \emph{Proceedings of the IEEE Conference on Computer Vision and
  Pattern Recognition (CVPR)}, 586--595.

\bibitem[{Zhao et~al.(2025)Zhao, Yang, Zhou, and Wang}]{Zhao2025RLIDM}
Zhao, B.; Yang, C.; Zhou, Q.; and Wang, Q. 2025.
\newblock RLI-DM: Robust Layout-Based Iterative Diffusion Model for SAR-to-RGB
  Image Translation.
\newblock \emph{IEEE Transactions on Geoscience and Remote Sensing}, 63:
  5108009.

\bibitem[{Zhao et~al.(2023)Zhao, Chen, Chen, Bao, Hao, Yuan, and
  Wong}]{Zhao2023UniControlNet}
Zhao, S.; Chen, D.; Chen, Y.-C.; Bao, J.; Hao, S.; Yuan, L.; and Wong, K.-Y.~K.
  2023.
\newblock Uni-ControlNet: All-in-One Control to Text-to-Image Diffusion Models.
\newblock In \emph{Advances in Neural Information Processing Systems
  (NeurIPS)}, volume~36, 11127--11150.

\bibitem[{Zheng et~al.(2025)Zheng, Ma, Tong, and Xie}]{Zheng2025RAE}
Zheng, B.; Ma, N.; Tong, S.; and Xie, S. 2025.
\newblock Diffusion Transformers with Representation Autoencoders.
\newblock arXiv:2510.11690.

\bibitem[{Zhu et~al.(2017{\natexlab{a}})Zhu, Park, Isola, and
  Efros}]{Zhu2017CycleGAN}
Zhu, J.-Y.; Park, T.; Isola, P.; and Efros, A.~A. 2017{\natexlab{a}}.
\newblock Unpaired Image-to-Image Translation Using Cycle-Consistent
  Adversarial Networks.
\newblock In \emph{Proceedings of the IEEE International Conference on Computer
  Vision (ICCV)}, 2223--2232.

\bibitem[{Zhu et~al.(2017{\natexlab{b}})Zhu, Zhang, Pathak, Darrell, Efros,
  Wang, and Shechtman}]{Zhu2017Multimodal}
Zhu, J.-Y.; Zhang, R.; Pathak, D.; Darrell, T.; Efros, A.~A.; Wang, O.; and
  Shechtman, E. 2017{\natexlab{b}}.
\newblock Toward Multimodal Image-to-Image Translation.
\newblock In \emph{Advances in Neural Information Processing Systems
  (NeurIPS)}, 465--476.

\end{thebibliography}


\begin{thebibliography}{19}
\providecommand{\natexlab}[1]{#1}

\bibitem[{Deng et~al.(2009)Deng, Dong, Socher, Li, Li, and
  Li}]{Deng2009ImageNet}
Deng, J.; Dong, W.; Socher, R.; Li, L.-J.; Li, K.; and Li, F.-F. 2009.
\newblock {ImageNet}: A Large-Scale Hierarchical Image Database.
\newblock In \emph{2009 IEEE Conference on Computer Vision and Pattern
  Recognition}, 248--255.

\bibitem[{Fan et~al.(2026)Fan, Diao, Wang, Lin, and Liu}]{Fan2026Prism}
Fan, W.; Diao, H.; Wang, Q.; Lin, D.; and Liu, Z. 2026.
\newblock The Prism Hypothesis: Harmonizing Semantic and Pixel Representations
  via Unified Autoencoding.
\newblock In \emph{Proceedings of the European Conference on Computer Vision
  (ECCV)}.

\bibitem[{Hu et~al.(2022)Hu, Shen, Wallis, Allen-Zhu, Li, Wang, Wang, and
  Chen}]{Hu2022LoRA}
Hu, E.~J.; Shen, Y.; Wallis, P.; Allen-Zhu, Z.; Li, Y.; Wang, S.; Wang, L.; and
  Chen, W. 2022.
\newblock {LoRA}: Low-Rank Adaptation of Large Language Models.
\newblock In \emph{International Conference on Learning Representations}.

\bibitem[{Huang et~al.(2021)Huang, Xu, Qian, Shi, Zhang, Bao, Wang, Liu, and
  Xiang}]{Huang2021QXSSAROPT}
Huang, M.; Xu, Y.; Qian, L.; Shi, W.; Zhang, Y.; Bao, W.; Wang, N.; Liu, X.;
  and Xiang, X. 2021.
\newblock The QXS-SAROPT Dataset for Deep Learning in SAR-Optical Data Fusion.
\newblock arXiv:2103.08259.

\bibitem[{Khanna et~al.(2024)Khanna, Liu, Zhou, Meng, Rombach, Burke, Lobell,
  and Ermon}]{Khanna2024DiffusionSat}
Khanna, S.; Liu, P.; Zhou, L.; Meng, C.; Rombach, R.; Burke, M.; Lobell, D.~B.;
  and Ermon, S. 2024.
\newblock {DiffusionSat}: A Generative Foundation Model for Satellite Imagery.
\newblock In \emph{Proceedings of the International Conference on Learning
  Representations (ICLR)}.

\bibitem[{Leng et~al.(2025)Leng, Singh, Hou, Xing, Xie, and
  Zheng}]{Leng2025REPAE}
Leng, X.; Singh, J.; Hou, Y.; Xing, Z.; Xie, S.; and Zheng, L. 2025.
\newblock {REPA-E}: Unlocking {VAE} for End-to-End Tuning of Latent Diffusion
  Transformers.
\newblock In \emph{Proceedings of the IEEE/CVF International Conference on
  Computer Vision (ICCV)}, 18262--18272.

\bibitem[{Liu et~al.(2025)Liu, Chen, Zhao, Zou, and Shi}]{Liu2025Text2Earth}
Liu, C.; Chen, K.; Zhao, R.; Zou, Z.; and Shi, Z. 2025.
\newblock {Text2Earth}: Unlocking Text-Driven Remote Sensing Image Generation
  with a Global-Scale Dataset and a Foundation Model.
\newblock \emph{IEEE Geoscience and Remote Sensing Magazine}, 13(3): 238--259.

\bibitem[{Naeem et~al.(2020)Naeem, Oh, Uh, Choi, and Yoo}]{Naeem2020Reliable}
Naeem, M.~F.; Oh, S.~J.; Uh, Y.; Choi, Y.; and Yoo, J. 2020.
\newblock Reliable Fidelity and Diversity Metrics for Generative Models.
\newblock In \emph{Proceedings of the 37th International Conference on Machine
  Learning}, volume 119 of \emph{Proceedings of Machine Learning Research},
  7176--7185.

\bibitem[{Oquab et~al.(2024)Oquab, Darcet, Moutakanni, Vo, Szafraniec,
  Khalidov, Fernandez, Haziza, Massa, El-Nouby, Assran, Ballas, Galuba, Howes,
  Huang, Li, Misra, Rabbat, Sharma, Synnaeve, Xu, J\'egou, Mairal, Labatut,
  Joulin, and Bojanowski}]{Oquab2024DINOv2}
Oquab, M.; Darcet, T.; Moutakanni, T.; Vo, H.; Szafraniec, M.; Khalidov, V.;
  Fernandez, P.; Haziza, D.; Massa, F.; El-Nouby, A.; Assran, M.; Ballas, N.;
  Galuba, W.; Howes, R.; Huang, P.-Y.; Li, S.-W.; Misra, I.; Rabbat, M.;
  Sharma, V.; Synnaeve, G.; Xu, H.; J\'egou, H.; Mairal, J.; Labatut, P.;
  Joulin, A.; and Bojanowski, P. 2024.
\newblock DINOv2: Learning Robust Visual Features without Supervision.
\newblock \emph{Transactions on Machine Learning Research}.

\bibitem[{Pan et~al.(2025)Pan, Lei, Fu, Li, Liu, Sun, He, Peng, Huang, and
  Zhao}]{Pan2025EarthSynth}
Pan, J.; Lei, S.; Fu, Y.; Li, J.; Liu, Y.; Sun, Y.; He, X.; Peng, L.; Huang,
  X.; and Zhao, B. 2025.
\newblock {EarthSynth}: Generating Informative Earth Observation with Diffusion
  Models.
\newblock arXiv:2505.12108.

\bibitem[{Robinson et~al.(2019)Robinson, Hou, Malkin, Soobitsky, Czawlytko,
  Dilkina, and Jojic}]{Robinson2019LandCover}
Robinson, C.; Hou, L.; Malkin, K.; Soobitsky, R.; Czawlytko, J.; Dilkina, B.;
  and Jojic, N. 2019.
\newblock Large Scale High-Resolution Land Cover Mapping with Multi-Resolution
  Data.
\newblock In \emph{Proceedings of the IEEE/CVF Conference on Computer Vision
  and Pattern Recognition (CVPR)}, 12726--12735.

\bibitem[{Shermeyer et~al.(2020)Shermeyer, Hogan, Brown, Van~Etten, Weir,
  Pacifici, H{\"a}nsch, Bastidas, Soenen, Bacastow, and
  Lewis}]{Shermeyer2020SpaceNet6}
Shermeyer, J.; Hogan, D.; Brown, J.; Van~Etten, A.; Weir, N.; Pacifici, F.;
  H{\"a}nsch, R.; Bastidas, A.; Soenen, S.; Bacastow, T.~M.; and Lewis, R.
  2020.
\newblock SpaceNet 6: Multi-Sensor All Weather Mapping Dataset.
\newblock In \emph{Proceedings of the IEEE/CVF Conference on Computer Vision
  and Pattern Recognition Workshops (CVPRW)}, 768--777.

\bibitem[{Sim\'eoni et~al.(2025)Sim\'eoni, Vo, Seitzer, Baldassarre, Oquab,
  Jose, Khalidov, Szafraniec, Yi, Ramamonjisoa, Massa, Haziza, Wehrstedt, Wang,
  Darcet, Moutakanni, Sentana, Roberts, Vedaldi, Tolan, Brandt, Couprie,
  Mairal, J\'egou, Labatut, and Bojanowski}]{Simeoni2025DINOv3}
Sim\'eoni, O.; Vo, H.~V.; Seitzer, M.; Baldassarre, F.; Oquab, M.; Jose, C.;
  Khalidov, V.; Szafraniec, M.; Yi, S.; Ramamonjisoa, M.; Massa, F.; Haziza,
  D.; Wehrstedt, L.; Wang, J.; Darcet, T.; Moutakanni, T.; Sentana, L.;
  Roberts, C.; Vedaldi, A.; Tolan, J.; Brandt, J.; Couprie, C.; Mairal, J.;
  J\'egou, H.; Labatut, P.; and Bojanowski, P. 2025.
\newblock DINOv3.
\newblock arXiv:2508.10104.

\bibitem[{Singh et~al.(2026)Singh, Zheng, Wu, Zhang, Shechtman, and
  Xie}]{Singh2026RAEv2}
Singh, J.; Zheng, B.; Wu, Z.; Zhang, R.; Shechtman, E.; and Xie, S. 2026.
\newblock Improved Baselines with Representation Autoencoders.
\newblock arXiv:2605.18324.

\bibitem[{Tang et~al.(2024)Tang, Cao, Hou, Jiang, Liu, and
  Meng}]{Tang2024CRSDiff}
Tang, D.; Cao, X.; Hou, X.; Jiang, Z.; Liu, J.; and Meng, D. 2024.
\newblock {CRS-Diff}: Controllable Remote Sensing Image Generation with
  Diffusion Model.
\newblock \emph{IEEE Transactions on Geoscience and Remote Sensing}, 62: 1--14.

\bibitem[{Yu et~al.(2025{\natexlab{a}})Yu, Kwak, Jang, Jeong, Huang, Shin, and
  Xie}]{Yu2025REPA}
Yu, S.; Kwak, S.; Jang, H.; Jeong, J.; Huang, J.; Shin, J.; and Xie, S.
  2025{\natexlab{a}}.
\newblock Representation Alignment for Generation: Training Diffusion
  Transformers Is Easier Than You Think.
\newblock In \emph{Proceedings of the International Conference on Learning
  Representations (ICLR)}.

\bibitem[{Yu et~al.(2025{\natexlab{b}})Yu, Liu, Liu, Shi, and
  Zou}]{Yu2025MetaEarth}
Yu, Z.; Liu, C.; Liu, L.; Shi, Z.; and Zou, Z. 2025{\natexlab{b}}.
\newblock {MetaEarth}: A Generative Foundation Model for Global-Scale Remote
  Sensing Image Generation.
\newblock \emph{IEEE Transactions on Pattern Analysis and Machine
  Intelligence}, 47(3): 1764--1781.

\bibitem[{Yue et~al.(2026)Yue, Hu, Chen, Zhang, Pan, Liu, Wang, Lan, Zhu,
  Zheng, and Wang}]{Yue2026PAE}
Yue, Z.; Hu, T.; Chen, M.; Zhang, H.; Pan, Z.; Liu, T.; Wang, Z.; Lan, J.; Zhu,
  X.; Zheng, B.; and Wang, Y. 2026.
\newblock What Matters for Diffusion-Friendly Latent Manifold? Prior-Aligned
  Autoencoders for Latent Diffusion.
\newblock arXiv:2605.07915.

\bibitem[{Zheng et~al.(2025)Zheng, Ma, Tong, and Xie}]{Zheng2025RAE}
Zheng, B.; Ma, N.; Tong, S.; and Xie, S. 2025.
\newblock Diffusion Transformers with Representation Autoencoders.
\newblock arXiv:2510.11690.

\end{thebibliography}

\end{document}


\twocolumn[
\vspace*{0.625in}
\begin{center}
{\LARGE\bfseries Supplementary Material}
\end{center}
\vspace{0.15in}
]
\appendix

\section{Theoretical Basis for Learning the Target Priors Before Image Translation}
\label{app:theory}

\subsection{Conditional Field and Population Risk Decomposition}
\label{app:risk-decomposition}

\paragraph{Assumptions and Notations.}
Let $X$ and $Y$ denote the source and target observations in a paired task, respectively, and let $Z=E(Y)$ be the target latent under a fixed encoder $E$. We use the Gaussian probability path
\[
Z_t=\alpha_t Z+\sigma_t\varepsilon,
\qquad
\varepsilon\sim\mathcal N(0,I),
\]
where $t$ and $\varepsilon$ are sampled independently of each other and of $(X,Y)$. We set $\omega=(Z_t,t)$ and use $\xi=\psi_t(Z,\varepsilon)$ as the flow matching prediction target.

Let $p^\star(x,y)$ be the joint density of the paired task and $p_t^\star(z_t,x)$ its induced perturbed density. Paired task data induce $p$ over $(\omega,X,\xi)$, while unpaired target images induce $q$ over $(\omega,\xi)$. Both use the same encoder, time sampling, probability path, and prediction parameterization. We assume that the relevant perturbed densities are positive and differentiable in $z_t$ and that $\xi$ is square integrable under $p$ and $q$.

\paragraph{Conditional Score Decomposition.}
According to Bayes' theorem,
\begin{equation}
p_t^\star(z_t\mid x)
=
\frac{p_t^\star(x\mid z_t)p_t^\star(z_t)}
{p^\star(x)}.
\label{eq:app-bayes}
\end{equation}
Taking the logarithm of Equation~\eqref{eq:app-bayes} and computing the gradient with respect to $z_t$ yields
\begin{equation}
\begin{aligned}
\nabla_{z_t}\log p_t^\star(z_t\mid x)
&=
\nabla_{z_t}\log p_t^\star(x\mid z_t)
+
\nabla_{z_t}\log p_t^\star(z_t)\\
&\quad-
\nabla_{z_t}\log p^\star(x)\\
&=
\nabla_{z_t}\log p_t^\star(z_t)
+
\nabla_{z_t}\log p_t^\star(x\mid z_t),
\end{aligned}
\label{eq:app-score-decomposition}
\end{equation}
The final equality follows because $p^\star(x)$ is independent of $z_t$. The first term is the marginal score of the task target distribution, determined by target images alone, whereas the second captures the dependence between source and target observations in paired data. For $\sigma_t>0$, the Gaussian perturbation identity gives
\begin{equation}
\mathbb E_p\!\left[
\varepsilon\mid Z_t=z_t,t,X=x
\right]
=
-\sigma_t\nabla_{z_t}\log p_t^\star(z_t\mid x).
\label{eq:app-noise-score}
\end{equation}

\paragraph{From Scores to Flow Matching Prediction Fields.}
Consider the general linear prediction target $\xi=\lambda_tZ+\mu_t\varepsilon$. For $\alpha_t\neq0$ and $\sigma_t>0$, it can be written as
\begin{equation}
\xi=
\frac{\lambda_t}{\alpha_t}Z_t+
\left(
\mu_t-\frac{\lambda_t\sigma_t}{\alpha_t}
\right)\varepsilon.
\label{eq:app-linear-target}
\end{equation}
Equation~\eqref{eq:app-noise-score} and its marginal form give
\begin{equation}
\begin{aligned}
F_p^\star(\omega,x)
&=
\mathbb E_p[\xi\mid\omega,X=x]
=
A_tz_t+
B_t\nabla_{z_t}\log p_t^\star(z_t\mid x),\\
U_p(\omega)
&=
\mathbb E_p[\xi\mid\omega]
=
A_tz_t+
B_t\nabla_{z_t}\log p_t^\star(z_t),
\end{aligned}
\end{equation}
where $A_t=\lambda_t/\alpha_t$ and $B_t=\sigma_t(\lambda_t\sigma_t/\alpha_t-\mu_t)$. Using Equation~\eqref{eq:app-score-decomposition}, we obtain
\begin{equation}
\begin{aligned}
C_p(\omega,x)
&:=
F_p^\star(\omega,x)-U_p(\omega)\\
&=
B_t\nabla_{z_t}\log p_t^\star(x\mid z_t),\\
F_p^\star(\omega,x)
&=
U_p(\omega)+C_p(\omega,x).
\end{aligned}
\label{eq:app-conditional-flow-component}
\end{equation}
The common term $A_tz_t$ cancels when the fields are subtracted, so $U_p$ represents the target marginal field and $C_p$ represents the source conditioned component in the flow matching prediction field.

\paragraph{Prediction Field Decomposition Relative to the Pretrained Prior.}
For any square integrable predictor $F(\omega,X)$, the population risk on the paired task is
\begin{equation}
\mathcal R_p(F)
=
\mathbb E_p\!\left[
\|\xi-F(\omega,X)\|_2^2
\right].
\label{eq:app-paired-risk}
\end{equation}
Since $F_p^\star(\omega,x)=\mathbb E_p[\xi\mid\omega,X=x]$, the projection property of conditional expectation gives
\begin{equation}
\mathcal R_p(F)-\mathcal R_p(F_p^\star)
=
\mathbb E_p\!\left[
\|F-F_p^\star\|_2^2
\right].
\label{eq:app-projection-risk}
\end{equation}
The definition of $C_p$ and the law of total expectation give
\begin{equation}
\mathbb E_p[C_p(\omega,X)\mid\omega]=0.
\label{eq:app-zero-mean}
\end{equation}
The prior training distribution $q$ induces
\begin{equation}
U_q(\omega)=\mathbb E_q[\xi\mid\omega].
\label{eq:app-prior-field}
\end{equation}
We assume that $U_q$ is defined and square integrable on the $\omega$ support under $p$. Using $F_p^\star=U_p+C_p$, the optimal task field can be written relative to the pretrained prior as
\begin{equation}
F_p^\star(\omega,x)
=
U_q(\omega)
+
\bigl(U_p(\omega)-U_q(\omega)\bigr)
+
C_p(\omega,x).
\label{eq:app-prior-decomposition}
\end{equation}
Writing $F=U_q+g$, Equation~\eqref{eq:app-projection-risk} gives
\begin{equation}
\mathcal R_p(U_q+g)-\mathcal R_p(F_p^\star)
=
\mathbb E_p\!\left[
\left\|
g-C_p-(U_p-U_q)
\right\|_2^2
\right].
\label{eq:app-prior-referenced-risk}
\end{equation}
The minimizing residual field is therefore
\begin{equation}
g^\star(\omega,x)
=
C_p(\omega,x)
+
\bigl(U_p(\omega)-U_q(\omega)\bigr).
\label{eq:app-fixed-prior}
\end{equation}
Since $U_p-U_q$ depends only on $\omega$, Equation~\eqref{eq:app-zero-mean} gives
\begin{equation}
\begin{aligned}
\mathbb E_p\!\left[
\left\langle C_p,U_p-U_q\right\rangle
\right]
&=0,\\
\mathbb E_p\!\left[
\|g^\star\|_2^2
\right]
&=
\mathbb E_p\!\left[
\|C_p\|_2^2
\right]
+
\mathbb E_p\!\left[
\|U_p-U_q\|_2^2
\right].
\end{aligned}
\label{eq:app-fixed-energy}
\end{equation}
For a fixed task $p$, closer agreement between $U_q$ and $U_p$ reduces the residual prior correction. Together, $C_p$ and $U_p-U_q$ define the population optimal modification $g^\star$ for paired adaptation under squared risk. This functional decomposition motivates P-DART to provide source conditioned control and backbone LoRA to provide adaptation capacity for the residual prior correction. The paired objective supervises the complete modification $g^\star$, and the two modules are jointly optimized within the coupled translation model.

\section{Implementation Details}
\label{app:implementation}

\subsection{Two-Stage Training Pipeline}
\label{app:training-config}

Current mainstream remote sensing generative models, including DiffusionSat, CRS-Diff, MetaEarth, Text2Earth, and EarthSynth, predominantly reuse the pretrained weights of Stable Diffusion or inherit the traditional diffusion architectures it represents \cite{Khanna2024DiffusionSat,Tang2024CRSDiff,Yu2025MetaEarth,Liu2025Text2Earth,Pan2025EarthSynth}. The improvements in these works primarily focus on remote sensing domain adaptation, conditional control, training data construction, and generation scale expansion, with less emphasis placed on the representation learning capabilities and training efficiency of the generative networks themselves. For downstream tasks, it is a common practice to directly adapt existing remote sensing generative models or initialize from pretrained diffusion weights derived from the natural image domain, aiming to reduce training costs and shorten experimental cycles. However, this paradigm of reusing weights inherently inherits the limitations of existing generative architectures and their respective latent spaces. Recent studies surrounding REPA, RAE, REPA-E, and PAE further demonstrate that the semantic organization, spatial structure, and local continuity of latent representations directly dictate the optimization difficulty, convergence efficiency, and ultimate generation performance of generative models \cite{Yu2025REPA,Zheng2025RAE,Leng2025REPAE,Yue2026PAE}.

Our objective is to independently learn a target-domain generative prior under a limited computational budget, while facilitating rapid iterations of model design and scaling experiments. To this end, we adopt RAEv2 as the foundational training framework, deliberately eschewing the reuse of generative weights from existing Stable Diffusion or contemporary remote sensing generative models \cite{Singh2026RAEv2}. RAE directly utilizes the structured representations extracted by a pretrained vision encoder as the generative latent space, enabling the diffusion model to learn on representations enriched with high-level semantics, thereby exhibiting excellent convergence efficiency and model scalability. However, the original RAE solely relies on the final-layer features of the encoder, failing to fully exploit the local textures and fine-grained spatial information preserved in the intermediate layers; consequently, it underperforms specially designed VAEs in reconstructing local structures and high-frequency details. Through parameter-free multi-layer feature aggregation, RAEv2 improves reconstruction quality without increasing the dimensionality of the latent representations, while concurrently preserving the semantic structure and generative learnability of the latent space. With the exception of the frozen vision encoder, the generative prior is trained entirely from random initialization. This design strikes a better balance among reconstruction fidelity, training efficiency, and generative learnability, enabling the independent pretraining and subsequent scaling experiments of the target-domain generative prior to be conducted efficiently within a constrained computational budget. Accordingly, we treat representation decoder fine-tuning as codec preparation. The subsequent LTP-BIT training pipeline comprises two decoupled stages, target-domain prior learning and paired conditional adaptation. The specific configurations are detailed below.

\paragraph{Codec Preparation.}

\begin{figure}[t]
\centering
\includegraphics[width=\columnwidth]{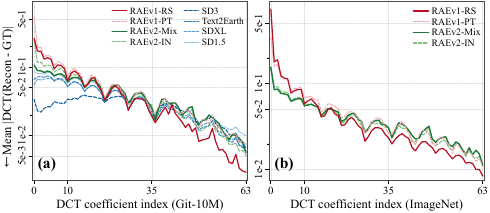}
\caption{Reconstruction errors across frequency components for RAE decoder variants. Following the frequency-domain analysis method in PRISM~\citep{Fan2026Prism}, we first average the RGB reconstruction residuals across the color channels, partition them into $8\times8$ non-overlapping patches, and compute the DCT patch by patch. The resulting 64 frequency coefficients are arranged in a zigzag order from the DC component to the high-frequency components. (a) Comparison among RAE-PT, RAE-RS, RAEv2-IN, RAEv2-Mix, and representative VAEs using 5K images randomly sampled from Git-10M. (b) Comparison of the four RAE decoders using 5K images from the ImageNet validation set.}
\label{fig:decoder-adaptation-spectrum}
\end{figure}

Before these two stages, we evaluate the direct domain fine-tuning of the RAE decoder. We denote the model loaded with the official weights pretrained on natural images as RAE-PT. Following the official training configuration, we fine-tune it for 20 epochs on one million images randomly sampled from Git-10M; the resulting model is denoted as RAE-RS. Figure~\ref{fig:decoder-adaptation-spectrum} illustrates the reconstruction errors in the Discrete Cosine Transform (DCT) domain on Git-10M and ImageNet~\citep{Deng2009ImageNet}, respectively, to compare the variations across different frequency components for both RAE and RAEv2 before and after fine-tuning.

\begin{figure*}[t]
\centering
\includegraphics[width=\textwidth]{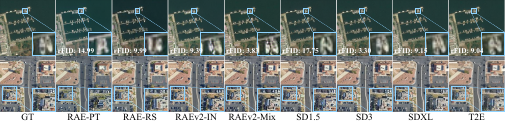}
\caption{Comparison of decoder reconstructions on remote sensing images. For two sets of Git-10M samples, the reconstruction results from left to right correspond to Ground Truth (GT), RAE-PT, RAE-RS, RAEv2-IN, RAEv2-Mix, SD1.5, SD3, SDXL, and Text2Earth VAE. The set-level FID, computed across 5K Git-10M evaluation images for each method, is annotated in the bottom-left corner of the first row (lower is better).}
\label{fig:decoder-adaptation-qualitative-rs}
\end{figure*}

\begin{figure}[t]
\centering
\includegraphics[width=\columnwidth]{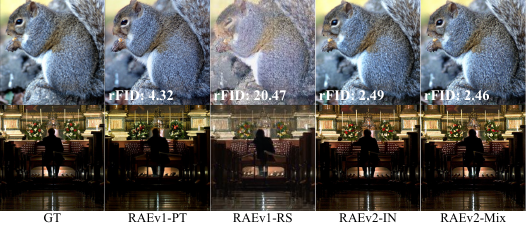}
\caption{Comparison of RAE decoder reconstructions on ImageNet. From left to right, the results correspond to Ground Truth (GT), RAE-PT, RAE-RS, RAEv2-IN, and RAEv2-Mix. The set-level FID, computed across 5K ImageNet validation images for each method, is annotated in the bottom-left corner of the first row (lower is better).}
\label{fig:decoder-adaptation-qualitative-imagenet}
\end{figure}

On Git-10M, RAE-RS achieves lower reconstruction errors in both the DC and high-frequency components compared to RAE-PT, with more pronounced improvements observed in the high-frequency band alongside a substantial reduction in FID. However, these quantitative improvements do not translate into perceptibly enhanced local details. As shown in Figure~\ref{fig:decoder-adaptation-qualitative-rs}, the reconstructions of ship hulls and pier boundaries by RAE-RS remain blurry. Vehicle contours, building edges, and roof textures exhibit no significant visual improvement over RAE-PT, remaining generally inferior to those produced by the SD-series VAEs. On ImageNet, although the reconstruction errors of RAE-RS in the mid- and high-frequency components decrease, the error in the DC component increases significantly. Both reconstruction examples in Figure~\ref{fig:decoder-adaptation-qualitative-imagenet} exhibit brightness and color shifts, coupled with lower contrast compared to RAE-PT. Furthermore, the fur texture of the squirrel and the fine-grained structures within the indoor scene are further degraded. These visual changes are consistent with the elevated DC component error observed in Figure~\ref{fig:decoder-adaptation-spectrum}(b).

To address the aforementioned deficiencies in local details and cross-domain degradation, we switch to RAEv2 and redesign the decoder training from the perspectives of both data coverage and optimization strategy. The training set comprises 1 million high-information-density remote sensing images filtered from Git-10M, an additional 1 million randomly sampled remote sensing images that do not overlap with the filtered subset, and 500,000 natural images sampled from ImageNet in a class-balanced manner. The high-information-density subset focuses on covering complex scenes enriched with urban building clusters and small objects, while the random subset serves to supplement the distributional diversity of remote sensing scenes. The ImageNet samples are included to preserve a unified codec for subsequent comparative analyses involving ImageNet generative pretraining. We denote the official ImageNet pretrained decoder as RAEv2-IN. Initialized with its weights, the model is trained on the aforementioned 2.5 million mixed images for 20 epochs, yielding a model we denote as RAEv2-Mix. The training process employs a strong-to-weak noise scheduling strategy for the latent features. Furthermore, the discriminator undergoes a warm-up phase for one epoch with the decoder frozen, followed by the joint optimization of pixel reconstruction, perceptual, and adversarial losses.

The results indicate that RAEv2-IN achieves reconstruction FIDs of 9.39 and 2.49 on Git-10M and ImageNet, respectively. Both values are lower than the 14.99 and 4.32 achieved by RAE-PT, demonstrating that RAEv2-IN serves as an overall stronger reconstruction baseline. Mixed-domain training further reduces the reconstruction FID on Git-10M to 3.83 and significantly lowers the reconstruction errors in the DC and low-frequency components. Correspondingly, small objects in the port, building edges, and roof textures in Figure~\ref{fig:decoder-adaptation-qualitative-rs} are reconstructed with enhanced clarity. On ImageNet, the frequency spectrum curves of RAEv2-Mix and RAEv2-IN nearly overlap, with reconstruction FIDs of 2.46 and 2.49, respectively. This indicates that the mixed-domain training successfully preserves the reconstruction capability for natural images. For all subsequent target-domain prior learning and paired conditional adaptation experiments, we uniformly freeze the RAEv2-Mix codec. This ensures that the performance disparities across different experimental groups stem primarily from these two learning stages.

\paragraph{Stage I. Target-Domain Prior Learning.}
In the first stage, we freeze the RAEv2-Mix codec and train the target-domain generative prior within the $16\times16\times1024$ latent space, employing the $\mathrm{DiT}^{\mathrm{DH}}$ architecture, which consists of a DiT encoding backbone and a shallow-and-wide DDT head~\citep{Zheng2025RAE}. The RAEv2 representation encoder performs parameter-free aggregation of features from the 11th, 13th, 15th, 17th, 19th, 21st, and 23rd layers of DINOv3-L, thereby fusing multi-level local structural and semantic information. The prior is trained by flow matching under $q_r$, the joint distribution over $(\omega,r,\xi)$ whose $(\omega,\xi)$ marginal is $q$ in Section~\ref{app:risk-decomposition}, and is conditioned exclusively on the nine-class spatial-resolution label $r$. An Internal Guidance (IG) auxiliary head in the intermediate backbone layers fits the same target as the full branch during training. Evaluation uses only the full branch, without IG extrapolation or Classifier-Free Guidance (CFG). Preliminary experiments showed no stable convergence gain from REPA while adding feature-extraction and representation-alignment overhead, so we omit it from formal training.

\paragraph{Stage II. Paired Conditional Adaptation.}
In the second stage, P-DART performs paired conditional adaptation through a source-conditioned reference stream and a generation stream, whose states are denoted by $c_t^j$ and $x_t^j$. The first P-DART location of each network region passes the reference state through unchanged. At each subsequent location, $c_t^j$ queries $[c_t^j;x_t^j]$, and $\operatorname{RefBlock}_{\phi_j}$ completes the residual and feed-forward update. The updated reference state supplies additional keys and values to generation attention. The learnable gate $b_j$ is added only to the logits of the appended reference-token keys, thereby controlling their value contribution. Reference tokens remain in their own stream, preserving the pretrained generation-token sequence. Each reference block is initialized from the corresponding prior layer by copying its first 25\% of attention heads and their Q, K, V, and O projections, and every gate logit is initialized to $-4$. We use $\phi$ for the reference-block and gate parameters and $\lambda$ for the LoRA parameters, which use standard zero-output initialization. With the pretrained base weights frozen, the complete P-DART reference stream and the LoRA parameters~\citep{Hu2022LoRA} in the designated generation-attention projections are jointly optimized under the paired flow-matching objective. Training uses no data augmentation.

We denote the P-DART locations in the DiT encoder and DDT head by $\mathcal I_E$ and $\mathcal I_D$, respectively. For P-DART Decoder, $\mathcal I_E=\emptyset$ and $\mathcal I_D$ contains all DDT-head layers. For P-DART Codec, $\mathcal I_E$ contains every other DiT backbone layer and $\mathcal I_D$ remains unchanged. Backbone LoRA uses the encoder locations, and the final reference-stream state enters the DDT head. The complete LTP-BIT training procedure is summarized in the accompanying algorithm.

\FloatBarrier
\begingroup
\setlength{\intextsep}{0pt}
\begin{algorithm}[H]
\caption*{\textbf{Algorithm}\quad LTP-BIT Training Procedure}
\fontsize{8}{9.2}\selectfont
\renewcommand{\algorithmicrequire}{\textbf{Input:}}
\renewcommand{\algorithmicensure}{\textbf{Output:}}
\begin{algorithmic}[1]
\newcommand{\ALGHEADING}[1]{%
  \item[]\hspace*{-\labelwidth}\hspace*{-\labelsep}\textbf{#1}}
\REQUIRE unpaired $q_r$, paired $p$, locations $\mathcal I_E,\mathcal I_D$;
$M$ updates with learning rates $\{\eta_m\}_{m=0}^{M-1}$
\ENSURE $\hat\theta,\phi^{(M)},\lambda^{(M)}$
\ALGHEADING{I. Target-domain prior learning}
\STATE $\displaystyle
\hat\theta\in\arg\min_{\theta}
\mathbb E_{(\omega,r,\xi)\sim q_r}
\bigl\|\xi-U_{\theta}(\omega,r)\bigr\|_2^2$
\ALGHEADING{II. Paired conditional adaptation}
\STATE $\displaystyle
\hat\theta\ \mathrm{fixed},\qquad
\phi^{(0)}\gets\operatorname{PriorInit}_{25\%}(\hat\theta)$
\STATE $\displaystyle
b_j^{(0)}\gets-4,\qquad j\in\mathcal I_E\cup\mathcal I_D$
\STATE $\displaystyle
\lambda^{(0)}\gets\operatorname{ZeroOutputInit}()$
\ALGHEADING{P-DART interaction at $j\in\mathcal I_E\cup\mathcal I_D$}
\STATE $\displaystyle
\widetilde c_t^j\gets\operatorname{Attn}_{\mathrm{ref}}
\left(
Q_c(c_t^j),K_c([c_t^j;x_t^j]),V_c([c_t^j;x_t^j])
\right),\quad j\ \text{not first}$
\STATE $\displaystyle
c_t^{j+1}\gets
\begin{cases}
c_t^j, & j\ \text{first},\\
\operatorname{RefBlock}_{\phi_j}(c_t^j,\widetilde c_t^j),
& j\ \text{not first}.
\end{cases}$
\STATE $\displaystyle
\widetilde x_t^j\gets\operatorname{Attn}_{g}
\left(
Q(x_t^j),K([x_t^j;c_t^{j+1}]),V([x_t^j;c_t^{j+1}]);b_j
\right)$
\FOR{$m=0,\ldots,M-1$}
    \STATE $\displaystyle
    (\omega,X,\xi)\sim p$
    \STATE $\displaystyle
    \widehat\xi_m\gets
    F_{\hat\theta,\phi^{(m)},\lambda^{(m)}}
    (\omega,X;\mathcal I_E,\mathcal I_D)$
    \STATE $\displaystyle
    \mathcal L_m\gets\bigl\|\xi-\widehat\xi_m\bigr\|_2^2$
\STATE $\displaystyle
    (\phi^{(m+1)},\lambda^{(m+1)})\gets
    (\phi^{(m)},\lambda^{(m)})-
    \eta_m\nabla_{\phi,\lambda}\mathcal L_m$
\ENDFOR
\end{algorithmic}
\end{algorithm}
\endgroup

\subsection{Prior Scaling Protocols}
\label{app:prior-scaling-protocols}

\begin{figure}[!t]
\centering
\includegraphics[width=\columnwidth]{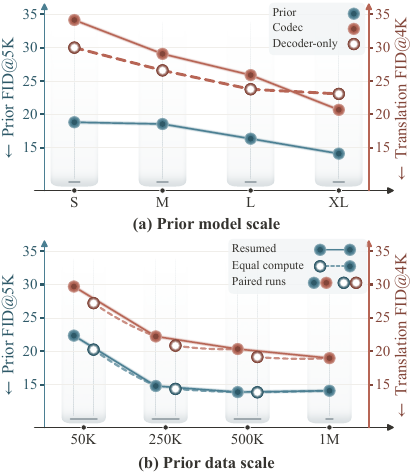}
\caption{Comprehensive evaluation of prior scaling and fixed-pool continued training controls. Blue and brownish-red represent the prior generation FID@5K and the translation FID@4K, respectively (lower is better). (a) Scaling up the prior model size while keeping the conditional capacity of the P-DART Decoder fixed; (b) Solid lines and solid markers indicate the continued training trajectory with progressively expanded data pools, whereas hollow markers denote the continued training controls initialized from the same parent exponential moving average (EMA) checkpoint while maintaining the original data pool. Dashed lines connect each hollow control point to the solid expansion point of the subsequent stage that shares an identical cumulative number of processed images. In the legend, ``Equal compute'' solely refers to the aforementioned matching of processed images, rather than implying strictly identical FLOPs or optimization trajectories; ``Paired runs'' indicates that a blue prior point and a brownish-red translation point with the same fill status originate from the same experimental group.}
\label{fig:prior-scaling-controls}
\end{figure}

For both types of scaling experiments, the prior training utilizes $256\times256$ inputs and spatial resolution conditions; during evaluation, the condition labels are sampled according to the spatial resolution distribution of the reference data.

\paragraph{Model-Capacity Scaling.}
Using the XL configuration of RAEv2 as a baseline, we construct four tiers of generative priors: S, M, L, and XL. Under the same $\mathrm{DiT}^{\mathrm{DH}}$ architecture, the S, M, and L variants systematically reduce the depth, width, and number of attention heads of the DiT encoding backbone, with the insertion depth of the auxiliary prediction head adjusted accordingly. Meanwhile, the DDT head, which consists of two layers, a hidden dimension of 2048, and 16 attention heads, remains constant across all four model configurations. All four generative priors are randomly initialized and independently trained on the same $\mathcal D_{1M}$ dataset for 15 epochs with a global batch size of 1024, employing the same optimization configuration and learning rate schedule. The comprehensive network configurations are detailed in Table~\ref{tab:model-scale-config}.

Each tier of the generative prior is adapted using identical QXS-SAROPT paired samples and training protocols. Serving as the primary trajectory for the model scaling experiments, the P-DART Decoder configuration maintains a fixed reference stream of 512 dimensions and 4 attention heads. Its encoder-side LoRA remains disabled, while the decoder-side LoRA rank is strictly fixed at 256. Since the DDT head remains invariant across all four prior tiers, the conditional structure and the number of trainable parameters under this configuration do not vary with the prior scale. The P-DART Codec serves as a supplementary control for the synchronous scaling of conditional capacity; its encoder-side reference module count, stream dimension, and LoRA rank scale proportionately alongside the DiT encoding backbone, whereas its decoder-side configuration remains unchanged. The prior architectures and conditional adaptation parameters for each model tier are detailed in Table~\ref{tab:model-scale-config}.

\begin{table*}[t]
\centering
\footnotesize
\setlength{\tabcolsep}{3.0pt}
\caption{Prior model scales and their paired conditional adaptation configurations. The four prior tiers employ the same pretraining data and training schedule. The P-DART Decoder maintains a constant conditional capacity, whereas the encoder-side reference stream of the P-DART Codec scales synchronously with the backbone prior.}
\label{tab:model-scale-config}
\begin{tabular}{@{}>{\raggedright\arraybackslash}p{0.30\textwidth}|*{4}{>{\centering\arraybackslash}p{0.16\textwidth}}@{}}
\specialrule{\heavyrulewidth}{0pt}{0pt}
\rowcolor{gray!12}
\multicolumn{1}{@{}>{\columncolor{gray!12}[0pt][\tabcolsep]}l|}{\rule[-1ex]{0pt}{3.4ex}Setting} &
S & M & L &
\multicolumn{1}{>{\columncolor{gray!12}[\tabcolsep][0pt]}c@{}}{XL} \\
\specialrule{0.05em}{0pt}{0pt}
\specialrule{0.05em}{1.2pt}{0pt}
\multicolumn{5}{@{}l@{}}{\rule[-0.8ex]{0pt}{3.0ex}\textcolor{black!60}{\usefont{T1}{ptm}{m}{it}Prior backbone}} \\
Parameters (M) & 204.6 & 282.4 & 423.4 & 873.8 \\
DiT backbone depth / width / heads & 12 / 512 / 8 & 16 / 768 / 12 & 20 / 1024 / 16 & 28 / 1440 / 20 \\
DDT head depth / width / heads & 2 / 2048 / 16 & 2 / 2048 / 16 & 2 / 2048 / 16 & 2 / 2048 / 16 \\
Auxiliary head attachment depth & 4 & 5 & 6 & 8 \\
MLP ratio & 4.0 & 4.0 & 4.0 & 4.0 \\
Latent grid / patch & $16^2$ / 1 & $16^2$ / 1 & $16^2$ / 1 & $16^2$ / 1 \\
Input channels & 1024 & 1024 & 1024 & 1024 \\
Condition tokens & $4+8$ & $4+8$ & $4+8$ & $4+8$ \\
\specialrule{\lightrulewidth}{1.2pt}{0pt}
\multicolumn{5}{@{}l@{}}{\rule[-0.8ex]{0pt}{3.0ex}\textcolor{black!60}{\usefont{T1}{ptm}{m}{it}Prior pretraining}} \\
Training images / epochs & 1M / 15 & 1M / 15 & 1M / 15 & 1M / 15 \\
Global batch size & 1024 & 1024 & 1024 & 1024 \\
Base / final LR & $2\times10^{-4}$ / $2\times10^{-5}$ & $2\times10^{-4}$ / $2\times10^{-5}$ & $2\times10^{-4}$ / $2\times10^{-5}$ & $2\times10^{-4}$ / $2\times10^{-5}$ \\
Warm-up / decay end & 5 / 10 & 5 / 10 & 5 / 10 & 5 / 10 \\
Input crop & $256^2$ & $256^2$ & $256^2$ & $256^2$ \\
Condition / dropout & Resolution / 0.1 & Resolution / 0.1 & Resolution / 0.1 & Resolution / 0.1 \\
\specialrule{\lightrulewidth}{1.2pt}{0pt}
\multicolumn{5}{@{}l@{}}{\rule[-0.8ex]{0pt}{3.0ex}\textcolor{black!60}{\usefont{T1}{ptm}{m}{it}Paired conditional adaptation}} \\
P-DART variants & Codec / Decoder & Codec / Decoder & Codec / Decoder & Codec / Decoder \\
Codec encoder reference blocks & 6 & 8 & 10 & 14 \\
Codec encoder stream width / heads & 128 / 2 & 192 / 3 & 256 / 4 & 360 / 5 \\
Encoder LoRA rank (Codec) & 48 & 64 & 96 & 128 \\
Encoder LoRA rank (Decoder) & 0 & 0 & 0 & 0 \\
Decoder reference stream width / heads & 512 / 4 & 512 / 4 & 512 / 4 & 512 / 4 \\
Decoder LoRA rank & 256 & 256 & 256 & 256 \\
Paired images / epochs & 16K / 80 & 16K / 80 & 16K / 80 & 16K / 80 \\
Global batch size & 256 & 256 & 256 & 256 \\
Peak LR & $5\times10^{-4}$ & $5\times10^{-4}$ & $5\times10^{-4}$ & $5\times10^{-4}$ \\
Noise curriculum & $1.0\ (10~\mathrm{ep})\rightarrow0.25$ & $1.0\ (10~\mathrm{ep})\rightarrow0.25$ & $1.0\ (10~\mathrm{ep})\rightarrow0.25$ & $1.0\ (10~\mathrm{ep})\rightarrow0.25$ \\
\bottomrule
\end{tabular}
\end{table*}

\paragraph{Prior data scaling.}
We freeze the RAEv2-Mix codec and the $\mathrm{DiT}^{\mathrm{DH}}$-L prior architecture, and construct strictly nested data pools $\mathcal D_{50K}\subset\mathcal D_{250K}\subset\mathcal D_{500K}\subset\mathcal D_{1M}$ from the same cleaned datasets. The initial prior is randomly initialized and independently trained on $\mathcal D_{50K}$. Subsequently, along the expansion trajectory, the EMA weights from the preceding stage are sequentially utilized to initialize the training on $\mathcal D_{250K}$, $\mathcal D_{500K}$, and $\mathcal D_{1M}$. For each continued training phase, the optimizer and the learning rate scheduler are re-initialized. Since the expansion trajectory simultaneously increases both data coverage and the cumulative number of processed images, we construct fixed-pool controls initialized from the same parent checkpoint, which maintain the original data pool while matching the number of newly processed training images. These controls match the parent checkpoint and the cumulative number of processed images, without requiring strictly identical FLOPs or optimization trajectories. The comprehensive initialization lineage and training schedules are detailed in Table~\ref{tab:data-scale-config}.

To investigate the impact of prior pretraining data, we freeze the $\mathrm{DiT}^{\mathrm{DH}}$-L and P-DART Codec configurations, and independently perform paired conditional adaptation on each prior checkpoint detailed in Table~\ref{tab:data-scale-config}. All experiments utilize the 16,000 training pairs from QXS-SAROPT, a global batch size of 256, and 5,040 training steps, while maintaining identical reference stream architectures, initialization schemes, optimization configurations, and noise schedules. Consequently, this set of experiments solely varies the target-domain prior loaded during conditional adaptation.

\begin{table*}[t]
\centering
\footnotesize
\setlength{\tabcolsep}{4.0pt}
\caption{Configurations for the Prior Data Scaling experiments. The expanded data pool branch and the fixed data pool branch are initialized from the same parent checkpoint and are matched by the cumulative number of processed images.}
\label{tab:data-scale-config}
\begin{tabular}{@{}ccccc|ccc@{}}
\specialrule{\heavyrulewidth}{0pt}{0pt}
\rowcolor{gray!12}
\multicolumn{1}{@{}>{\columncolor{gray!12}[0pt][\tabcolsep]}c}{\rule[-0.75ex]{0pt}{3.15ex}} &
& & & Cumulative images &
    \multicolumn{3}{|>{\columncolor{gray!12}[\tabcolsep][0pt]}c@{}}{LR schedule (epochs)} \\
\cline{6-8}
\rowcolor{gray!12}
\multicolumn{1}{@{}>{\columncolor{gray!12}[0pt][\tabcolsep]}c}{\multirow{-2}{*}{Prior images}} &
\multirow{-2}{*}{Training strategy} &
\multirow{-2}{*}{Initialization} &
\multirow{-2}{*}{Epochs} &
\rule[-0.75ex]{0pt}{3.15ex}seen &
Warm-up & Decay start &
\multicolumn{1}{>{\columncolor{gray!12}[\tabcolsep][0pt]}c@{}}{Decay end} \\
\specialrule{0.05em}{0pt}{0pt}
\specialrule{0.05em}{1.2pt}{0pt}
\rule{0pt}{2.6ex}50K  & From scratch           & Random          & 300 & 15M  & 100 & {--} & 200 \\
\midrule
250K & Expanded pool          & 50K checkpoint  & 100 & 40M  & 5  & 30 & 60 \\
50K  & Fixed-pool control     & 50K checkpoint  & 500 & 40M  & 5  & 30 & 60 \\
\midrule
500K & Expanded pool          & 250K checkpoint & 100 & 90M  & 5  & 30 & 60 \\
250K & Fixed-pool control     & 250K checkpoint & 200 & 90M  & 5  & 30 & 60 \\
\midrule
1M   & Expanded pool          & 500K checkpoint & 100 & 190M & 5  & 30 & 60 \\
500K & Fixed-pool control     & 500K checkpoint & 200 & 190M & 5  & 30 & 60 \\
\bottomrule
\end{tabular}
\par\vspace{2pt}
\scriptsize\textit{Note.} The cumulative number of processed images includes repeated accesses to the same image across epochs. Continued training inherits only the EMA model weights, while both the optimizer and the learning rate scheduler are re-initialized. All experiments employ a global batch size of 1024, with the base and final learning rates set to $2\times10^{-4}$ and $2\times10^{-5}$, respectively.
\end{table*}

\paragraph{Fixed-Pool Controls Matched by Cumulative Processed Images.}
Figure~\ref{fig:prior-scaling-controls} and Table~\ref{tab:data-scale-matched-exposure-results} comprehensively report the fixed-pool control experiments matched by the cumulative number of processed images. Figure~\ref{fig:prior-scaling-controls}(a) provides a reference for prior model scaling under a fixed conditional capacity; Figure~\ref{fig:prior-scaling-controls}(b) compares the expanded data pool branch with the fixed data pool branch. Both branches originate from the same parent EMA checkpoint and are evaluated at the milestones of 40M, 90M, and 190M cumulatively processed images. The area of the data points in the figure does not convey any quantitative semantics.

\begin{table*}[t]
\centering
\caption{
Prior data scaling results for the expanded data trajectory and fixed pool
continued training controls at matched cumulative training volumes. Each row
reports the parent checkpoint, training route, current data pool, prior
quality, and paired translation performance.
}
\label{tab:data-scale-matched-exposure-results}
\footnotesize
\setlength{\tabcolsep}{3.0pt}
\renewcommand{\arraystretch}{1.08}
\begin{tabular}{@{}ccll|cc|ccccc@{}}
\specialrule{\heavyrulewidth}{0pt}{0pt}
\rowcolor{gray!12}
\multicolumn{1}{
    @{}>{\columncolor{gray!12}[0pt][\tabcolsep]}c
}{
    \rule[-0.75ex]{0pt}{3.15ex}
} &
& & &
\multicolumn{2}{|c|}{Prior} &
\multicolumn{5}{
    >{\columncolor{gray!12}[\tabcolsep][0pt]}c@{}
}{
    Translation
} \\
\cline{5-6}
\cline{7-11}
\rowcolor{gray!12}
\multicolumn{1}{
    @{}>{\columncolor{gray!12}[0pt][\tabcolsep]}c
}{
    \multirow{-2}{*}{Cumulative images}
} &
\multirow{-2}{*}{Init. from} &
\multirow{-2}{*}{Branch} &
\multirow{-2}{*}{Pool} &
FID$\downarrow$ &
CMMD$\downarrow$ &
PSNR$\uparrow$ &
SSIM$\uparrow$ &
LPIPS$\downarrow$ &
FID$\downarrow$ &
\multicolumn{1}{
    >{\columncolor{gray!12}[\tabcolsep][0pt]}c@{}
}{
    CMMD$\downarrow$
} \\
\specialrule{0.05em}{0pt}{0pt}
\specialrule{0.05em}{1.2pt}{0pt}
\rule{0pt}{2.6ex}15M
& Random & Initial & 50K
& 22.35 & 0.856
& 16.093 & 0.373 & 0.469 & 29.70 & 0.253 \\
\midrule
40M
& 50K & Expanded & 250K
& 14.81 & 0.794
& 16.009 & 0.361 & 0.460 & 22.23 & 0.231 \\
40M
& 50K & Fixed & 50K
& 20.25 & 0.833
& 16.057 & 0.369 & 0.466 & 27.21 & 0.241 \\
\midrule
90M
& 250K & Expanded & 500K
& 13.89 & 0.773
& 15.912 & 0.356 & 0.459 & 20.34 & 0.226 \\
90M
& 250K & Fixed & 250K
& 14.37 & 0.782
& 15.946 & 0.359 & 0.460 & 20.86 & 0.222 \\
\midrule
190M
& 500K & Expanded & 1M
& 14.09 & 0.768
& 15.845 & 0.352 & 0.461 & 18.98 & 0.224 \\
190M
& 500K & Fixed & 500K
& 13.86 & 0.762
& 15.805 & 0.353 & 0.460 & 19.15 & 0.220 \\
\bottomrule
\end{tabular}
\end{table*}

Overall, the results exhibit a clear scaling trend. As the target-domain data pool and cumulative training volume increase, both the unconditional prior and the corresponding translation model achieve progressively stronger distribution-level performance. Along the expanded-pool trajectory, prior FID decreases from 22.35 to approximately 14, prior CMMD decreases from 0.856 to 0.768, translation FID decreases from 29.70 to 18.98, and translation CMMD decreases from 0.253 to 0.224. The fixed-pool controls follow a similar but generally slower trajectory, showing that continued optimization also improves distribution modeling, while incorporating additional target-domain images accelerates this process at early and intermediate training stages. In comparison, PSNR, SSIM, and LPIPS vary within relatively narrow ranges, indicating that prior-data scaling primarily strengthens distributional realism rather than paired reconstruction fidelity. The gains gradually saturate as the cumulative training volume increases.

\subsection{Baseline Implementations and Fair Comparison Protocol}
\label{app:baseline-config}

For all baseline methods, we follow the official implementations and training recipes, while aligning the number of training epochs and learning rate settings with our experiments. This gives all methods the same paired data exposure, while their computational costs and optimization trajectories remain specific to their respective implementations. The network architecture, initialization, objective functions, optimizer, data augmentation, and remaining hyperparameters follow the official settings of each method. During paired conditional adaptation, all methods use the same dataset splits, preprocessing procedures, input resolution, and evaluation protocol.

\subsection{Pretrained Model Implementations}
\label{app:model-config}

\paragraph{Prior Models for Method Analysis.} The model-scale studies in Table~1(a) and (b) of the main text compare S, M, L, and XL priors independently pretrained on RS-1M for 15 epochs. The pretraining-data-scale study in Table~1(c) and Table~2 fixes the L-scale prior architecture and varies the strictly nested pretraining pools from 50K to 1M. The target-domain enrichment study in Table~3 fixes the total pool size at 50K and varies the number of QXS target images; all variants are initialized from the same L-scale prior trained on 50K images for 300 epochs and are subsequently trained for 100 additional epochs.

\paragraph{Prior Model for Baseline Comparisons.} For the principal quantitative and qualitative comparisons with existing methods, our model uses the $\mathrm{DiT}^{\mathrm{DH}}$-XL prior pretrained for 100 epochs. Its base corpus, RS-1M, comprises one million remote sensing RGB images. Following the findings in Supplementary Section~\ref{app:prior-to-translation}, we add target modality images from the training splits of QXS-SAROPT, SpaceNet6, and Chesapeake to RS-1M for prior pretraining.

\paragraph{Prior Models for Ablation Studies.} The prior initialization ablation in Table~\ref{tab:prior-initialization} compares random initialization and ImageNet initialization with the $\mathrm{DiT}^{\mathrm{DH}}$-XL prior pretrained on RS-1M for 15 epochs. All variants undergo paired conditional adaptation for 80 epochs with the same paired training data, optimization settings, and evaluation protocol. Experiments that explicitly vary prior initialization, model scale, or pretraining configuration use the corresponding prior specified in the relevant experiment.

\section{Dataset Construction}
\label{app:data}

\subsection{Unpaired Target-Domain Data Construction}
\label{app:unpaired-target-data-construction}

This section describes the source data, preprocessing, quality control,
resolution stratification, and nested data pool construction used to prepare
the unpaired RGB corpus for target-domain prior pretraining.

\paragraph{Git-10M.}
Git-10M is a global-scale remote sensing image-text dataset containing more
than ten million image-text pairs with geographic and spatial-resolution
metadata~\citep{Liu2025Text2Earth}. The dataset covers diverse geographic
regions and land-surface scenes at multiple spatial resolutions. We use its
RGB imagery and Ground Sample Distance (GSD) metadata as the candidate source
for constructing the target-domain pretraining corpus.

\paragraph{RS-1M Construction.}
Starting from the RGB imagery in Git-10M, we construct RS-1M through a
resolution-aware curation pipeline that integrates standardized cropping,
quality screening, redundancy suppression, and diversity-aware sampling.
Each image is center-cropped into a $256\times256$ patch and assigned to one
of nine GSD categories at $0.5$, $1$, $2$, $4$, $8$, $16$, $32$, $64$, or
$128\,\mathrm{m/pixel}$. We allocate a fixed sampling quota to each category,
as summarized in Table~\ref{tab:unpaired-target-resolution}, to maintain broad
coverage across spatial resolutions.

The initial screening removes patches affected by invalid padding, severe
brightness clipping, blur, or occlusion. We subsequently suppress redundancy
using dHash to identify duplicate and near-duplicate images, followed by
DINOv2-based nearest-neighbor analysis to identify candidates with high visual
similarity~\citep{Oquab2024DINOv2}. Within each GSD category, we estimate image
information density from local entropy, spatial frequency, edge density, and
local intensity variation. These measurements increase the inclusion
probability of structurally informative samples, while similarity constraints
limit the concentration of visually repetitive content and preserve coverage
across diverse land-surface scenes.

A final quality review examines the selected images for residual blur, cloud
cover, and occlusion. The resulting RS-1M inventory contains one million RGB
patches with associated GSD category labels. From this fixed inventory, we
construct strictly nested subsets
$\mathcal D_{50K}\subset\mathcal D_{250K}\subset
\mathcal D_{500K}\subset\mathcal D_{1M}$.
This nesting preserves sample identity across scales and provides the data
pools used in the target-domain prior scaling experiments.

\begin{table}[t]
\centering
\footnotesize
\setlength{\tabcolsep}{12pt}
\renewcommand{\arraystretch}{1.12}
\caption{Spatial-resolution distribution of the constructed RS-1M dataset.}
\label{tab:unpaired-target-resolution}
\begin{tabular}{@{}lc@{}}
\specialrule{\heavyrulewidth}{0pt}{0pt}
\rowcolor{gray!12}
\multicolumn{1}{
    @{}>{\columncolor{gray!12}[0pt][\tabcolsep]}l
}{
    \rule[-0.75ex]{0pt}{3.15ex}Resolution
} &
\multicolumn{1}{
    >{\columncolor{gray!12}[\tabcolsep][0pt]}c@{}
}{
    Number of Images
} \\
\specialrule{0.05em}{0pt}{0pt}
\specialrule{0.05em}{1.2pt}{0pt}
\rule{0pt}{2.6ex}$0.5\,\mathrm{m/pixel}$ & 344,669 \\
$1\,\mathrm{m/pixel}$   & 206,801 \\
$2\,\mathrm{m/pixel}$   & 124,081 \\
$4\,\mathrm{m/pixel}$   & 74,449 \\
$8$--$128\,\mathrm{m/pixel}$ &
\begin{tabular}[c]{@{}c@{}}
50,000\\[-2pt]
per resolution
\end{tabular} \\
\midrule
Total & 1,000,000 \\
\bottomrule
\end{tabular}
\end{table}

\subsection{Paired Translation Benchmark Construction}
\label{app:dataset-construction}

This section details the partitioning, cropping, and cleaning procedures for the three benchmarks utilized in the main text. Unless otherwise specified, all randomized data splitting and sampling procedures use a fixed random seed of 42. All datasets employ $256\times256$ paired patches as the fundamental units for training and evaluation, with both modalities sharing identical spatial windows and file identifiers. Upon completion of data construction, we verify the pairing relationships, image dimensions, and sample quantities.

\paragraph{QXS-SAROPT.}
QXS-SAROPT (QXS) comprises 20,000 pairs of single-polarization SAR and RGB patches from San Diego, Shanghai, and Qingdao~\citep{Huang2021QXSSAROPT}. The SAR images were acquired by the Gaofen-3 C-band sensor in spotlight mode, while the RGB images were sourced from Google Earth. The patches have dimensions of $256\times256$ pixels and a spatial resolution of $1\,\mathrm{m/pixel}$. We apply a randomized $4{:}1$ split to the entire dataset and freeze the resulting filename inventory across all experiments, ultimately yielding 16,000 training pairs and 4,000 testing pairs.

\paragraph{SpaceNet6.}
SpaceNet6 consists of SAR and RGB imagery covering the Rotterdam area~\citep{Shermeyer2020SpaceNet6}, both featuring a spatial resolution of $0.5\,\mathrm{m/pixel}$. The SAR images were captured by Capella Space and contain four polarization channels: HH, HV, VH, and VV. The RGB images are derived from the Maxar WorldView-2 PS-RGB product. We partition all 3,401 pairs of $900\times900$ tiles into training and test sets at a fixed ratio of $4{:}1$, yielding 2,720 and 681 tiles, respectively.

We independently normalize the four polarization channels using the 1st and 99th percentiles of each SAR tile, and subsequently composite HH, $(\mathrm{HV}+\mathrm{VH})/2$, and VV into a three-channel input. Following this, we synchronously crop $4\times4$ SAR and RGB patches using four starting coordinates, $[0, 215, 429, 644]$, along both the horizontal and vertical axes. This results in an average uniaxial overlap rate of 16.15\% between adjacent windows. This cropping strategy yields 43,520 training pairs and 10,896 test pairs. After removing RGB patches whose fill fraction or border-fill fraction exceeds the predefined threshold of 0.1\%, we retain 20,168 training pairs and 5,048 test pairs.

\paragraph{Chesapeake.}
We construct a NIR$\rightarrow$RGB benchmark based on the CVPR 2019 Chesapeake Land Cover dataset~\citep{Robinson2019LandCover}. Covering six U.S. states, this dataset includes four-band NAIP imagery with a spatial resolution of $1\,\mathrm{m/pixel}$, land cover labels, and building masks. We merge the official training and validation partitions to form the training candidate pool and retain the official test partition as the test candidate pool. Exact duplicate tiles across the six states are resolved before cropping, with test-set membership taking precedence. We then extract non-overlapping $256\times256$ patches using a stride of 256 pixels. The first three bands serve as the RGB target, while the fourth band acts as the NIR input, which is duplicated across three channels to form the input tensor.

Within each candidate pool, samples are selected through mutually exclusive semantic, random, and information-density branches at a ratio of $8{:}1{:}1$. Semantic selection uses the land cover labels and building masks, whereas the random and information-density branches draw from the remaining candidates. This produces final training and test sets at a ratio of $4{:}1$, comprising 16,000 and 4,000 pairs, respectively.

\section{Mechanism Validation Details}
\label{app:mechanism-validation}

\subsection{Local Feature-Structure Compatibility Analysis}
\label{app:local-feature-compatibility}

To validate the local feature-structure compatibility hypothesis proposed in the main text, we design a single-step target feature recovery experiment. Under identical target images and perturbation states, this experiment independently measures the one-step target feature recovery rates of the frozen target prior and the corresponding conditional translation model, serving to investigate the correspondence between the local recovery capability of the prior and the translation-side recovery capability following the introduction of the SAR condition.

Given the $i$-th paired sample $(x_i,y_i)$, where $x_i$ and $y_i$ denote the SAR condition and the corresponding target-domain image, respectively, we first utilize the frozen encoder and decoder to construct the target latent variable and the reference image

\begin{equation}
z_{i,0}=E(y_i),
\qquad
y_{i,\mathrm{ref}}=D(z_{i,0}),
\end{equation}

where $E$ and $D$ denote the encoder and decoder, respectively. Utilizing the encoded-decoded result as the reference ensures that all feature distances are computed within the same codec representation space, thereby preventing the codec's inherent reconstruction bias from being factored into the recovery error.

Subsequently, for each perturbation timestep $t$, we construct the perturbed state of the target latent variable $z_{i,0}$ according to the linear probability path:
\begin{equation}
z_{i,t}
=
(1-t)z_{i,0}
+
t\varepsilon_{i,t},
\qquad
\varepsilon_{i,t}\sim\mathcal{N}(0,I).
\end{equation}
For the prior-side one-step recovery, the frozen target prior estimates the clean latent without SAR conditioning or guidance as
\begin{equation}
\hat z_{i,0}^{\mathrm{P}}
=
z_{i,t}
-
t\,v_{\theta}(z_{i,t},t;\emptyset).
\end{equation}
For the translation-side recovery, we provide the corresponding conditional translation model with the paired SAR observation $x_i$, and perform a one-step estimation under the same target latent variable, perturbation timestep, and noise realization:
\begin{equation}
\hat z_{i,0}^{\mathrm{T}}
=
z_{i,t}
-
t\,v_{\theta,\psi}(z_{i,t},t;x_i),
\end{equation}
where $\psi$ denotes the adaptation parameters within the conditional side chain and the prior backbone. The perturbed state and the two recovered states are decoded respectively as
\begin{equation}
y_{i,t}^{\mathrm{pre}}=D(z_{i,t}),
\qquad
y_{i,t}^{\mathrm{P}}=D(\hat z_{i,0}^{\mathrm{P}}),
\qquad
y_{i,t}^{\mathrm{T}}=D(\hat z_{i,0}^{\mathrm{T}}).
\end{equation}

We utilize a frozen DINOv3-SAT~\citep{Simeoni2025DINOv3} to extract spatially aligned patch features. Let the feature of the $p$-th patch be denoted by $\phi_p(\cdot)$; the average cosine distance prior to recovery is defined as
\begin{equation}
d_{i,t}^{\mathrm{pre}}
=
\frac{1}{P}
\sum_{p=1}^{P}
\left[
1-
\frac{
\phi_p(y_{i,\mathrm{ref}})^\top
\phi_p(y_{i,t}^{\mathrm{pre}})
}{
\|\phi_p(y_{i,\mathrm{ref}})\|_2
\|\phi_p(y_{i,t}^{\mathrm{pre}})\|_2
}
\right].
\end{equation}
For recovery mode $m\in\{\mathrm{P},\mathrm{T}\}$, the average cosine distance post-recovery is given by
\begin{equation}
d_{i,t}^{\mathrm{post},m}
=
\frac{1}{P}
\sum_{p=1}^{P}
\left[
1-
\frac{
\phi_p(y_{i,\mathrm{ref}})^\top
\phi_p(y_{i,t}^{m})
}{
\|\phi_p(y_{i,\mathrm{ref}})\|_2
\|\phi_p(y_{i,t}^{m})\|_2
}
\right].
\end{equation}
The corresponding Target Feature Recovery Ratio (TFRR) is defined as
\begin{equation}
\mathrm{TFRR}_{i,t}^{m}
=
\frac{
d_{i,t}^{\mathrm{pre}}
-
d_{i,t}^{\mathrm{post},m}
}{
d_{i,t}^{\mathrm{pre}}
}
\times 100\%.
\end{equation}
where $d_{i,t}^{\mathrm{pre}}>0$. $\mathrm{TFRR}^{\mathrm{P}}$ and $\mathrm{TFRR}^{\mathrm{T}}$ correspond to the Prior TFRR and Translation TFRR referenced in the main text, respectively. A higher TFRR signifies that the one-step recovery eliminates a greater extent of the target feature deviation introduced by the perturbation; $\mathrm{TFRR}=0$ indicates that the feature distance remains unchanged post-recovery, whereas a negative value implies an increase in feature deviation following recovery.

To unify the perturbation x-axis across different target samples, we utilize the median pre-recovery distance of all target samples at the selected terminal timestep of the schedule, $t_{\star}=0.9801$, as a shared reference:
\begin{equation}
c_{i,t}
=
\frac{
d_{i,t}^{\mathrm{pre}}
}{
\operatorname{median}_{j}
\left(d_{j,t_{\star}}^{\mathrm{pre}}\right)
}
\times 100\%.
\end{equation}

Here, $c_{i,t}$ corresponds to the normalized feature perturbation degree shown in the figures. This normalization is solely employed to determine the x-axis coordinates and perturbation bins; it is not involved in the computation of the TFRR. A value of $100\%$ corresponds to the median distance across all target samples at the selected terminal timestep; consequently, the $c_{i,t}$ for an individual state can exceed $100\%$.

The main analysis includes four $\mathrm{DiT}^{\mathrm{DH}}$-L priors trained on 50K, 250K, 500K, and 1M target-domain data, alongside their corresponding conditional translation models. On the translation side, SAR conditional control and the prior backbone LoRA are enabled, without employing class conditioning or guidance enhancement. We evaluate all models on the 4,000 paired samples from the QXS test set using 50 fixed perturbation timesteps within $[0.05,0.995]$. All priors and their corresponding conditional translation models share the same target samples, perturbation timesteps, and noise realizations.

We pool the normalized pre-recovery distances of the 4,000 target images across the 50 states and establish 50 shared quantile bins based on their overall distribution. The same bin boundaries are applied to all priors and both recovery modes. Let $\mathcal{S}_{i,b}$ denote the set of perturbation states falling into the $b$-th bin that belong to image $i$, and let $\mathcal{I}_b$ denote the set of images containing valid states. The recovery rate reported in the figure is given by
\begin{equation}
\overline{\mathrm{TFRR}}_{b}^{m}
=
\frac{1}{|\mathcal{I}_b|}
\sum_{i\in\mathcal{I}_b}
\left[
\frac{1}{|\mathcal{S}_{i,b}|}
\sum_{t\in\mathcal{S}_{i,b}}
\mathrm{TFRR}_{i,t}^{m}
\right].
\end{equation}

This aggregation is first averaged within the same target image and subsequently averaged across different images, ensuring that each target image carries equal weight; no smoothing is applied to any of the curves. To prevent numerical amplification of the TFRR under low perturbations, the main text exclusively reports bins with an average pre-recovery distance of no less than $0.05$.

Panels (c)--(f) of the local feature-structure compatibility analysis figure in the main text pair the Prior TFRR and Translation TFRR under identical prior scales and perturbation bins. The Pearson correlation coefficients are computed based on all paired bins with an average pre-recovery distance of no less than $0.05$, without any additional filtering based on the recovery outcomes. The pooled correlation coefficient aggregates the valid bins across the four prior scales, whereas the correlation coefficients within each individual scale are computed from the valid bins within their respective groups. All reported $r$ values represent descriptive Pearson correlation coefficients at the bin level.

\subsection{Target Feature Manifold Coverage Analysis}
\label{app:target-manifold-coverage}

To validate the target feature manifold coverage hypothesis proposed in the main text, we compute Coverage and Density~\citep{Naeem2020Reliable} under a unified generation and feature evaluation protocol. For each prior to be evaluated, we generate $M=5{,}000$ target-domain images of size $256\times256$ in the absence of SAR conditioning and guidance enhancement. All models consistently utilize the $1\,\mathrm{m/pixel}$ resolution category, 50 sampling steps, and a random seed of 42, with both CFG and IG set to 1. The real reference comprises all $N=4{,}000$ optical images from the QXS test set. All priors share the same real reference, number of generated samples, and sampling configurations.

We utilize a frozen DINOv2 ViT-L/14~\citep{Oquab2024DINOv2} to extract the CLS features of the images, and subsequently apply $\ell_2$ normalization to the obtained features. Let the sets of normalized features for the real images and the prior-generated images be denoted respectively as
\begin{equation}
\mathcal{R}
=
\left\{
r_i
\right\}_{i=1}^{N},
\qquad
\mathcal{G}
=
\left\{
g_j
\right\}_{j=1}^{M}.
\end{equation}
For each real feature $r_i$, let $r_i^{(k)}$ denote the $k$-th nearest neighbor among the remaining real features, and define the local radius by the corresponding Euclidean distance
\begin{equation}
\rho_i^{(k)}
=
\left\|
r_i-r_i^{(k)}
\right\|_2.
\end{equation}
In the normalized feature space, Coverage is defined as the proportion of real local neighborhoods containing at least one generated feature
\begin{equation}
\operatorname{Coverage}_{k}
=
\frac{1}{N}
\sum_{i=1}^{N}
\mathbf{1}
\left[
\min_{1\leq j\leq M}
\left\|
r_i-g_j
\right\|_2
\leq
\rho_i^{(k)}
\right].
\end{equation}
Density is defined as the average number of overlaps where generated features fall into real local neighborhoods, normalized by $k$
\begin{equation}
\operatorname{Density}_{k}
=
\frac{1}{kM}
\sum_{j=1}^{M}
\sum_{i=1}^{N}
\mathbf{1}
\left[
\left\|
r_i-g_j
\right\|_2
\leq
\rho_i^{(k)}
\right].
\end{equation}

Coverage characterizes the proportion of real local neighborhoods that the generated distribution can reach, with its value ranging within $[0,1]$; Density characterizes the degree of aggregation and overlap of generated features within these neighborhoods, and its value is not bounded by 1. The main text reports Coverage as a percentage and Density as a raw value. The local radius is determined by the same real reference and remains invariant across different priors.

The pretraining data scaling experiments employ the same $\mathrm{DiT}^{\mathrm{DH}}$-L architecture and evaluate four priors obtained from the strictly nested data pools of 50K, 250K, 500K, and 1M. The 50K prior is trained for 300 epochs, and each subsequent stage continues training for 100 epochs based on the weights from the preceding stage. The fixed-pool continued training controls, matched by the cumulative number of processed images, adopt the same evaluation protocol and are aggregated alongside the four primary trajectories of the expanded data pools.

\begin{table}[t]
\centering
\caption{Target manifold structure and paired translation performance of the fixed-pool continued training controls. The cumulative numbers of processed images for the 50K, 250K, and 500K groups are 40M, 90M, and 190M, respectively.}
\label{tab:fixed-pool-manifold-controls}
\scriptsize
\renewcommand{\arraystretch}{1.08}
\setlength{\tabcolsep}{2.3pt}
\resizebox{\columnwidth}{!}{%
\begin{tabular}{@{}l|cc|ccc@{}}
\specialrule{\heavyrulewidth}{0pt}{0pt}
\rowcolor{gray!12}
\multicolumn{1}{
    @{}>{\columncolor{gray!12}[0pt][\tabcolsep]}l|
}{
    \rule[-0.75ex]{0pt}{3.15ex}
} &
\multicolumn{2}{c|}{
    Manifold metrics
} &
\multicolumn{3}{
    >{\columncolor{gray!12}[\tabcolsep][0pt]}c@{}
}{
    Translation metrics
} \\
\cline{2-6}
\rowcolor{gray!12}
\multicolumn{1}{
    @{}>{\columncolor{gray!12}[0pt][\tabcolsep]}l|
}{
    \multirow{-2}{*}{Pool}
} &
Cov.@3 (\%)$\uparrow$ &
Dens.@3$\uparrow$ &
FID$\downarrow$ &
CMMD$\downarrow$ &
\multicolumn{1}{
    >{\columncolor{gray!12}[\tabcolsep][0pt]}c@{}
}{
    PSNR$\uparrow$
} \\
\specialrule{0.05em}{0pt}{0pt}
\specialrule{0.05em}{1.2pt}{0pt}
\rule{0pt}{2.6ex}50K
& 8.025 & 0.071 & 27.21 & 0.241 & 16.057 \\
250K
& 8.200 & 0.058 & 20.86 & 0.222 & 15.946 \\
500K
& 8.350 & 0.059 & 19.15 & 0.220 & 15.805 \\
\bottomrule
\end{tabular}
}
\end{table}

Table~\ref{tab:fixed-pool-manifold-controls} independently summarizes the results of the fixed-pool continued training. This group, together with the expanded data pool experiment of the subsequent tier, constitutes a comparison matched by the cumulative number of processed images; its manifold metrics are independently aggregated along the fixed-pool continued training trajectory.

The target-domain data enrichment experiments are all initialized with the same $\mathrm{DiT}^{\mathrm{DH}}$-L prior trained on 50K images for 300 epochs. The total amount of continued training data for each group is fixed at 50K, with equal numbers of samples from the original $\mathcal D_{50K}$ pool progressively replaced by strictly nested QXS subsets. This configuration results in target-domain sample scales of 4K, 8K, and 16K participating in the manifold comparisons, respectively; all settings undergo continued training for an additional 100 epochs. The resulting priors utilize the aforementioned unified generation and feature evaluation protocol to compute Coverage and Density. The corresponding conditional translation models utilize the same 16K QXS paired data, $\mathrm{DiT}^{\mathrm{DH}}$-L backbone, P-DART conditional side chain, and backbone LoRA, while maintaining consistent training budgets and evaluation protocols. We establish additional adaptation controls by freezing the prior backbone and exclusively training the P-DART conditional side chain, which serves to disentangle the respective impacts of the prior manifold structure and the backbone adaptation strategy on the translation results.

\begin{table}[t]
\centering
\caption{Target-domain data enrichment controls with a frozen prior backbone. Coverage and Density are computed from the corresponding priors, while only the P-DART conditional side chain is trained during paired conditional adaptation.}
\label{tab:frozen-backbone-enrichment-control}
\scriptsize
\renewcommand{\arraystretch}{1.08}
\setlength{\tabcolsep}{2.3pt}
\resizebox{\columnwidth}{!}{%
\begin{tabular}{@{}l|cc|ccc@{}}
\specialrule{\heavyrulewidth}{0pt}{0pt}
\rowcolor{gray!12}
\multicolumn{1}{
    @{}>{\columncolor{gray!12}[0pt][\tabcolsep]}l|
}{
    \rule[-0.75ex]{0pt}{3.15ex}
} &
\multicolumn{2}{c|}{
    Manifold metrics
} &
\multicolumn{3}{
    >{\columncolor{gray!12}[\tabcolsep][0pt]}c@{}
}{
    Translation metrics
} \\
\cline{2-6}
\rowcolor{gray!12}
\multicolumn{1}{
    @{}>{\columncolor{gray!12}[0pt][\tabcolsep]}l|
}{
    \multirow{-2}{*}{QXS}
} &
Cov.@3 (\%)$\uparrow$ &
Dens.@3$\uparrow$ &
FID$\downarrow$ &
CMMD$\downarrow$ &
\multicolumn{1}{
    >{\columncolor{gray!12}[\tabcolsep][0pt]}c@{}
}{
    PSNR$\uparrow$
} \\
\specialrule{0.05em}{0pt}{0pt}
\specialrule{0.05em}{1.2pt}{0pt}
\rule{0pt}{2.6ex}4K
& 21.275 & 0.132 & 24.33 & 0.241 & 14.748 \\
8K
& 29.075 & 0.203 & 23.76 & 0.234 & 14.726 \\
16K
& 36.900 & 0.294 & 23.81 & 0.228 & 14.677 \\
\bottomrule
\end{tabular}
}
\end{table}

Table~\ref{tab:frozen-backbone-enrichment-control} demonstrates that, when exclusively updating the conditional side chain, the breadth of target manifold coverage and the degree of local overlap continue to increase alongside the QXS sample scale. The translation distribution-level metrics exhibit overall improvement, while the instance-level metrics remain largely stable. This control, in conjunction with the backbone LoRA results presented in the main text, jointly indicates that the prior manifold structure dictates the range of accessible target representations, whereas the backbone adaptation strategy further influences how these representations are utilized for specific paired instances.

The absolute value of Coverage is simultaneously influenced by the number of real samples, the number of generated samples, and the neighborhood scale; therefore, all cross-comparisons in this paper are strictly constrained to a fixed $N=4{,}000$, $M=5{,}000$, and a shared feature extractor. Both Coverage and Density reported herein are computed from 5,000 prior-generated samples. The pretraining scaling experiments and the target-domain data enrichment experiments are analyzed independently, maintaining identical sampling parameters, random seeds, and real references within each respective experiment.

\subsection{From Target-Domain Priors to Cross-Modal Translation}
\label{app:prior-to-translation}

The analyses above reveal three connected patterns. (a) Prior TFRR and Translation TFRR rise together across pretraining scales and remain highly correlated, indicating that target-feature recovery learned by the prior is reflected in downstream translation. (b) Greater coverage and density of task-relevant target features are accompanied by better distribution-level translation performance, showing that prior utility depends on its alignment with the downstream target domain. (c) The theoretical decomposition identifies source conditioned control and residual prior correction as two components of paired adaptation. Guided by this functional view, we compare two conditional adaptation configurations across prior scales. The Decoder places P-DART and LoRA in the fixed DDT head, whereas the Codec scales the P-DART reference stream and backbone LoRA in the DiT backbone. Across the evaluated configurations, the Codec achieves a better balance between target realism and instance fidelity at the XL scale (Figure~\ref{fig:prior-scaling-controls}(a) and Table~1(b) of the main text).

Together, these observations motivate our choices for prior construction and conditional adaptation. We therefore incorporate each benchmark's training targets as unpaired samples during prior pretraining to strengthen task-relevant coverage, and scale the conditional pathway with the prior backbone so that richer target-domain representations can be translated into source-consistent outputs. Together, these choices reduce the divergence between distribution-level realism and instance-level fidelity.

\section{Ablation Studies}
\label{app:ablation}


\begin{table}[t]
\centering
\caption{Effect of backbone generative prior initialization on paired translation performance. We compare random initialization with generative pretraining on ImageNet and RS-1M.}
\label{tab:prior-initialization}
\scriptsize
\renewcommand{\arraystretch}{1.08}
\setlength{\tabcolsep}{3pt}
\resizebox{\columnwidth}{!}{%
\begin{tabular}{@{}l|ccccc@{}}
\specialrule{\heavyrulewidth}{0pt}{0pt}
\rowcolor{gray!12}
\multicolumn{1}{
    @{}>{\columncolor{gray!12}[0pt][\tabcolsep]}l|
}{
    \rule[-0.75ex]{0pt}{3.15ex}Prior
} &
PSNR$\uparrow$ &
SSIM$\uparrow$ &
LPIPS$\downarrow$ &
FID$\downarrow$ &
\multicolumn{1}{
    >{\columncolor{gray!12}[\tabcolsep][0pt]}c@{}
}{
    CMMD$\downarrow$
} \\
\specialrule{0.05em}{0pt}{0pt}
\specialrule{0.05em}{1.2pt}{0pt}
\rule{0pt}{2.6ex}Random
& 14.566
& \textbf{0.425}
& 0.829
& 453.73
& 6.270 \\
ImageNet
& 16.1782
& 0.3658
& 0.4514
& 23.6509
& 0.2466 \\
RS-1M
& \textbf{16.2187}
& 0.3613
& \textbf{0.4390}
& \textbf{19.9728}
& \textbf{0.2273} \\
\bottomrule
\end{tabular}
}
\end{table}

\subsection{Effect of Backbone Prior Initialization}

Table~\ref{tab:prior-initialization} compares random initialization, ImageNet pretraining, and the RS-1M XL prior pretrained for 15 epochs under the same paired adaptation protocol for 80 epochs on QXS. Random initialization yields the highest SSIM but performs substantially worse on the other metrics. Relative to ImageNet pretraining, the RS-1M prior improves PSNR by 0.0405 dB and reduces FID, CMMD, and LPIPS by 3.6781, 0.0193, and 0.0124, while SSIM decreases by 0.0045. These results indicate that the remote sensing prior improves perceptual and distributional quality while maintaining comparable reconstruction fidelity.

\begin{figure}[t]
\centering
\includegraphics[width=\columnwidth]{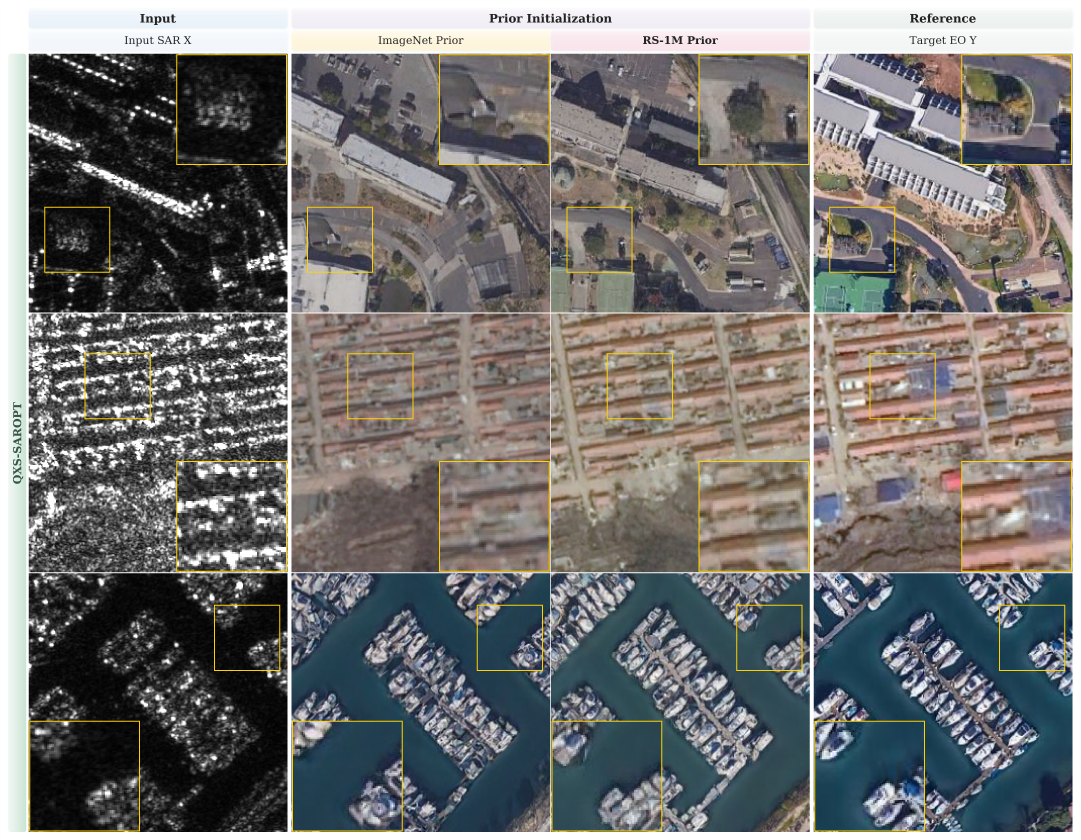}
\caption{Qualitative comparison of ImageNet and RS-1M prior initialization on QXS.}
\label{fig:prior_initialization_ablation}
\end{figure}

Figure~\ref{fig:prior_initialization_ablation} qualitatively compares ImageNet and RS-1M prior initialization on QXS. Both priors recover the overall scene layout, while the RS-1M prior better preserves building arrangements, vessel locations, and local boundaries and produces a more coherent optical appearance. This agrees with its higher PSNR and lower LPIPS, FID, and CMMD in Table~\ref{tab:prior-initialization}, while ImageNet initialization retains a slight SSIM advantage.

\begin{table}[t]
\centering
\caption{Quantitative comparison of different curriculum endpoints on
translation performance. The pure-noise replacement probability is kept at
100\% for the first 10 epochs, followed by a linear decay to 25\%.}
\label{tab:curriculum_endpoint}
\scriptsize
\renewcommand{\arraystretch}{1.08}
\setlength{\tabcolsep}{2.5pt}
\resizebox{\columnwidth}{!}{%
\begin{tabular}{@{}lc|ccccc@{}}
\specialrule{\heavyrulewidth}{0pt}{0pt}
\rowcolor{gray!12}
\multicolumn{1}{
    @{}>{\columncolor{gray!12}[0pt][\tabcolsep]}l
}{
    \rule[-0.75ex]{0pt}{3.15ex}Dataset
} &
End Epoch &
PSNR$\uparrow$ &
SSIM$\uparrow$ &
LPIPS$\downarrow$ &
FID$\downarrow$ &
\multicolumn{1}{
    >{\columncolor{gray!12}[\tabcolsep][0pt]}c@{}
}{
    CMMD$\downarrow$
} \\
\specialrule{0.05em}{0pt}{0pt}
\specialrule{0.05em}{1.2pt}{0pt}
\rule{0pt}{2.6ex}\multirow{2}{*}{Chesapeake}
& 80
& 17.5835
& 0.3463
& 0.2935
& 18.5616
& 0.8395 \\
& \textbf{20}
& \textbf{17.7171}
& \textbf{0.3500}
& \textbf{0.2904}
& \textbf{18.1939}
& \textbf{0.8247} \\
\midrule
\multirow{2}{*}{SpaceNet6}
& 80
& \textbf{18.9241}
& 0.3511
& 0.2962
& 44.2808
& 0.8327 \\
& \textbf{20}
& 18.9237
& \textbf{0.3526}
& \textbf{0.2926}
& \textbf{43.7616}
& \textbf{0.8084} \\
\midrule
\multirow{2}{*}{QXS}
& \textbf{80}
& \textbf{16.0406}
& \textbf{0.3518}
& \textbf{0.4441}
& 17.1970
& 0.2046 \\
& 20
& 15.9555
& 0.3505
& 0.4457
& \textbf{16.5363}
& \textbf{0.2009} \\
\bottomrule
\end{tabular}
}
\end{table}

\subsection{Effect of Curriculum Endpoint}

We compare two settings in which the decay phase concludes at the 20th and
80th epochs, respectively. As shown in
Table~\ref{tab:curriculum_endpoint}, the 20-epoch setting yields
comprehensive improvements across all metrics on the Chesapeake dataset. On
the SpaceNet6 dataset, apart from achieving a nearly identical PSNR, the
20-epoch setting outperforms the 80-epoch alternative across all other
metrics. On the QXS dataset, the 80-epoch setting achieves better
PSNR, SSIM, and LPIPS; however, the 20-epoch setting reduces the FID from
17.1970 to 16.5363 and the CMMD from 0.2046 to 0.2009, demonstrating a
trade-off between paired reconstruction quality and distributional
consistency.

Overall, concluding the curriculum earlier reserves more training stages for the joint learning of source observations and real optical targets, which consistently improves FID and CMMD across all three datasets. Based on the quantitative results, we ultimately set the curriculum endpoint to the 20th epoch.

\subsection{Effect of Fine-Tuning Strategy}

\begin{figure*}[p]
\centering
\captionsetup{justification=centering,singlelinecheck=false,skip=3pt}
\includegraphics[width=\textwidth,height=0.92\textheight,keepaspectratio,
trim=3bp 3bp 3bp 3bp,clip]{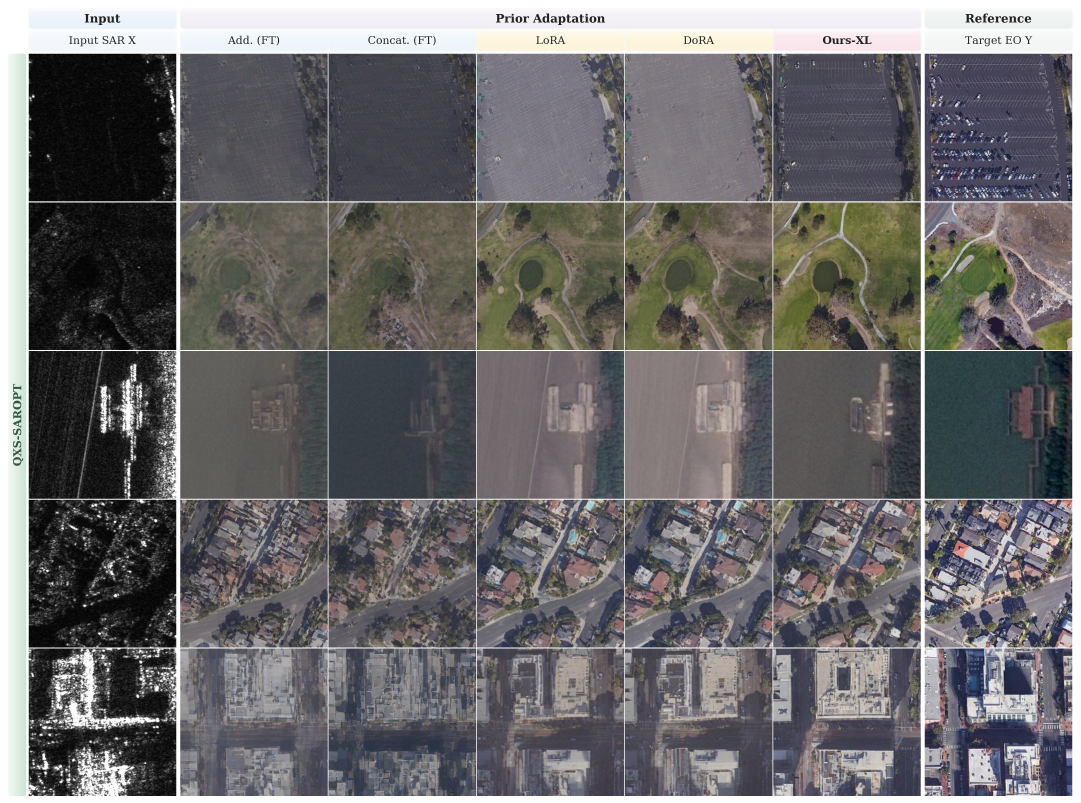}
\caption{Qualitative comparison of the fine-tuning strategies evaluated in
the main text on the QXS dataset.}
\label{fig:result_qxs2}
\end{figure*}

Figure~\ref{fig:result_qxs2} provides a visual comparison of the fine tuning strategies evaluated in Table~6 of the main text. Addition and concatenation with full fine tuning recover less faithful structures than LoRA and DoRA. Although LoRA and DoRA better preserve the overall scene layout, they still blur or omit fine structures, whereas Ours-XL more accurately reconstructs road geometry and parking lot markings. These results support the effectiveness of our joint adaptation design in source conditioned control.

\FloatBarrier

\section{Qualitative Results and Analysis}
\label{app:qualitative}

This section provides additional visualizations of the predictions generated by our proposed model and the baseline methods across the three datasets. These three benchmarks can be viewed as heterogeneous remote sensing image translation tasks with different cross-modal challenges. QXS and SpaceNet6 involve SAR-to-optical translation, where microwave backscattering responses do not have a deterministic correspondence with optical colors and textures and are further affected by speckle noise, layover, and complex scattering. The model must therefore preserve reliable spatial structures, such as buildings, roads, shorelines, ships, and field boundaries, while using the target-domain prior to synthesize plausible optical appearances. Chesapeake instead involves NIR-to-RGB translation. Although its source and target images are geometrically well aligned, a single NIR band does not uniquely determine visible-spectrum colors, requiring the model to recover the spectral differences among vegetation, water, bare land, and built-up areas. As shown in the qualitative comparisons, our method consistently balances source-structure preservation and target-domain realism across all three datasets. It better maintains building layouts, road orientations, ship locations, and shoreline contours while producing natural optical textures on QXS and SpaceNet6; on Chesapeake, it recovers plausible colors for fields, vegetation, and built-up regions while retaining clear boundaries and local details. In contrast, several competing methods exhibit structural mismatches, incorrectly generated content, blurred textures, or noticeable color shifts, demonstrating the stronger and more stable cross-modal translation capability of our method.

\begin{figure*}[p]
\centering
\captionsetup{justification=centering,singlelinecheck=false,skip=3pt}
\includegraphics[width=\textwidth,height=0.92\textheight,keepaspectratio,
trim=7bp 27bp 7bp 27bp,clip]{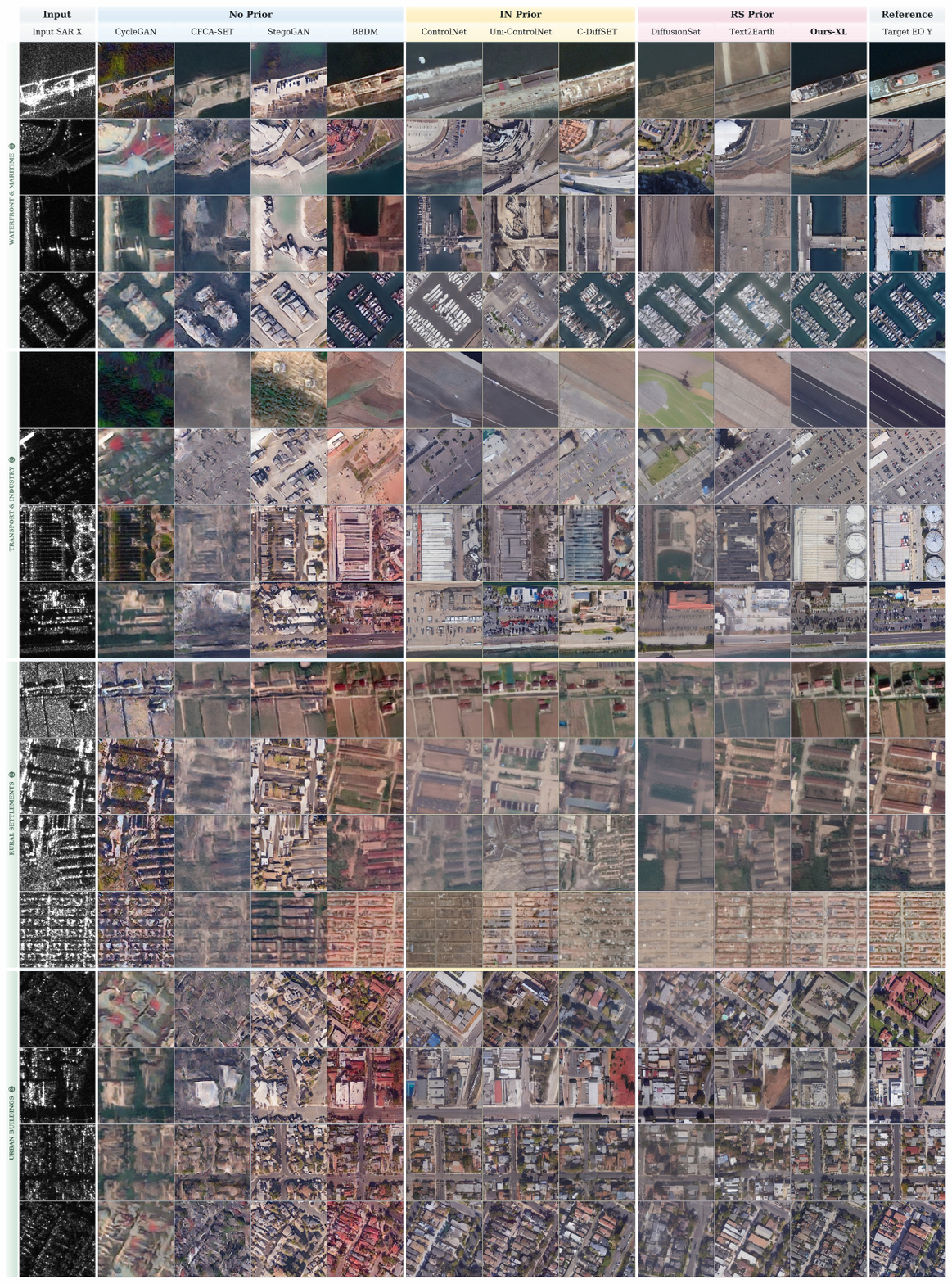}
\caption{Qualitative comparisons on the QXS dataset.}
\label{fig:result_qxs}
\end{figure*}

\begin{figure*}[p]
\centering
\captionsetup{justification=centering,singlelinecheck=false,skip=3pt}
\includegraphics[width=\textwidth,height=0.92\textheight,keepaspectratio,
trim=7bp 27bp 7bp 27bp,clip]{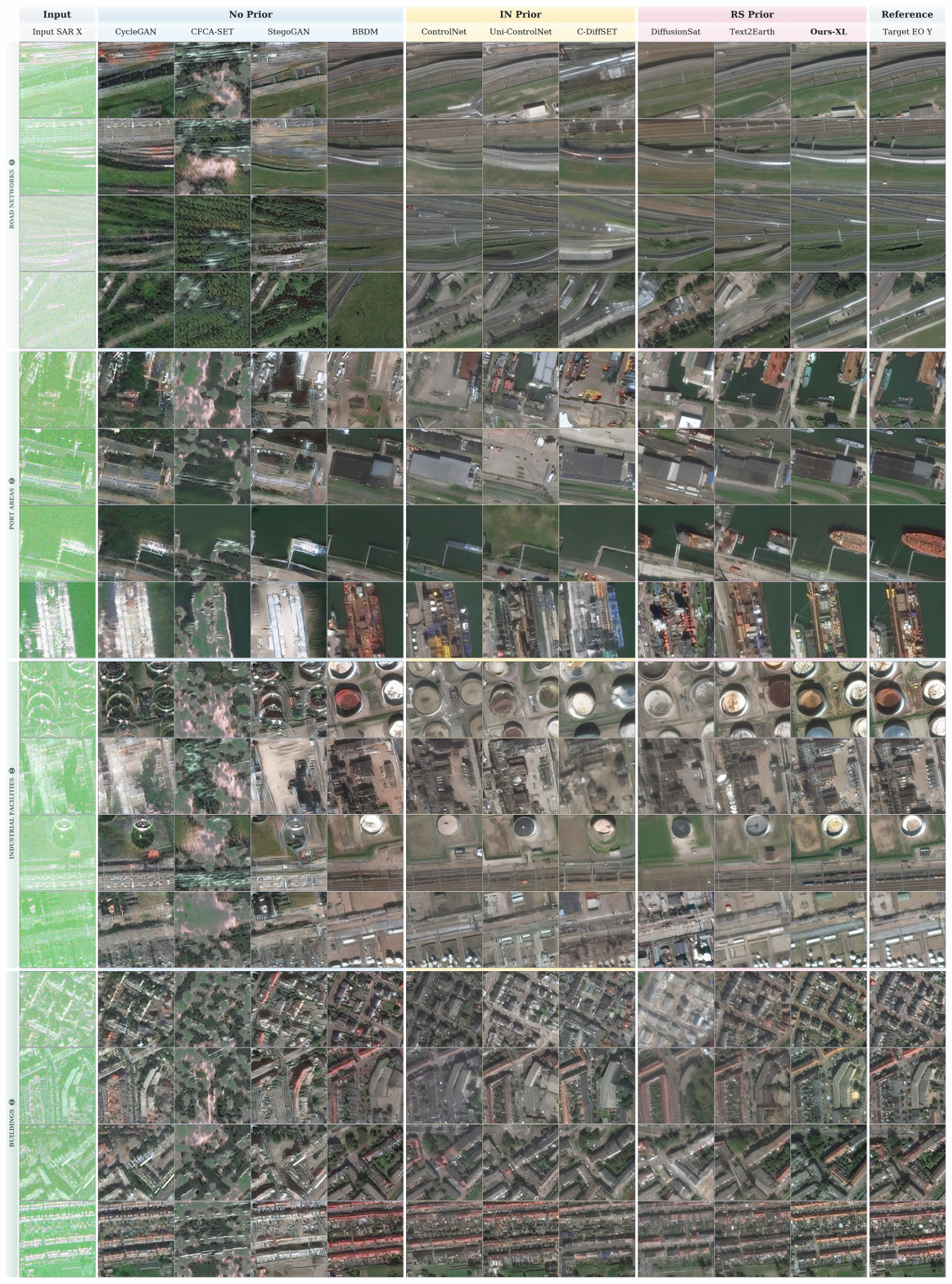}
\caption{Qualitative comparisons on the SpaceNet6 dataset.}
\label{fig:result_spacenet6}
\end{figure*}

\begin{figure*}[p]
\centering
\captionsetup{justification=centering,singlelinecheck=false,skip=3pt}
\includegraphics[width=\textwidth,height=0.92\textheight,keepaspectratio,
trim=7bp 27bp 7bp 27bp,clip]{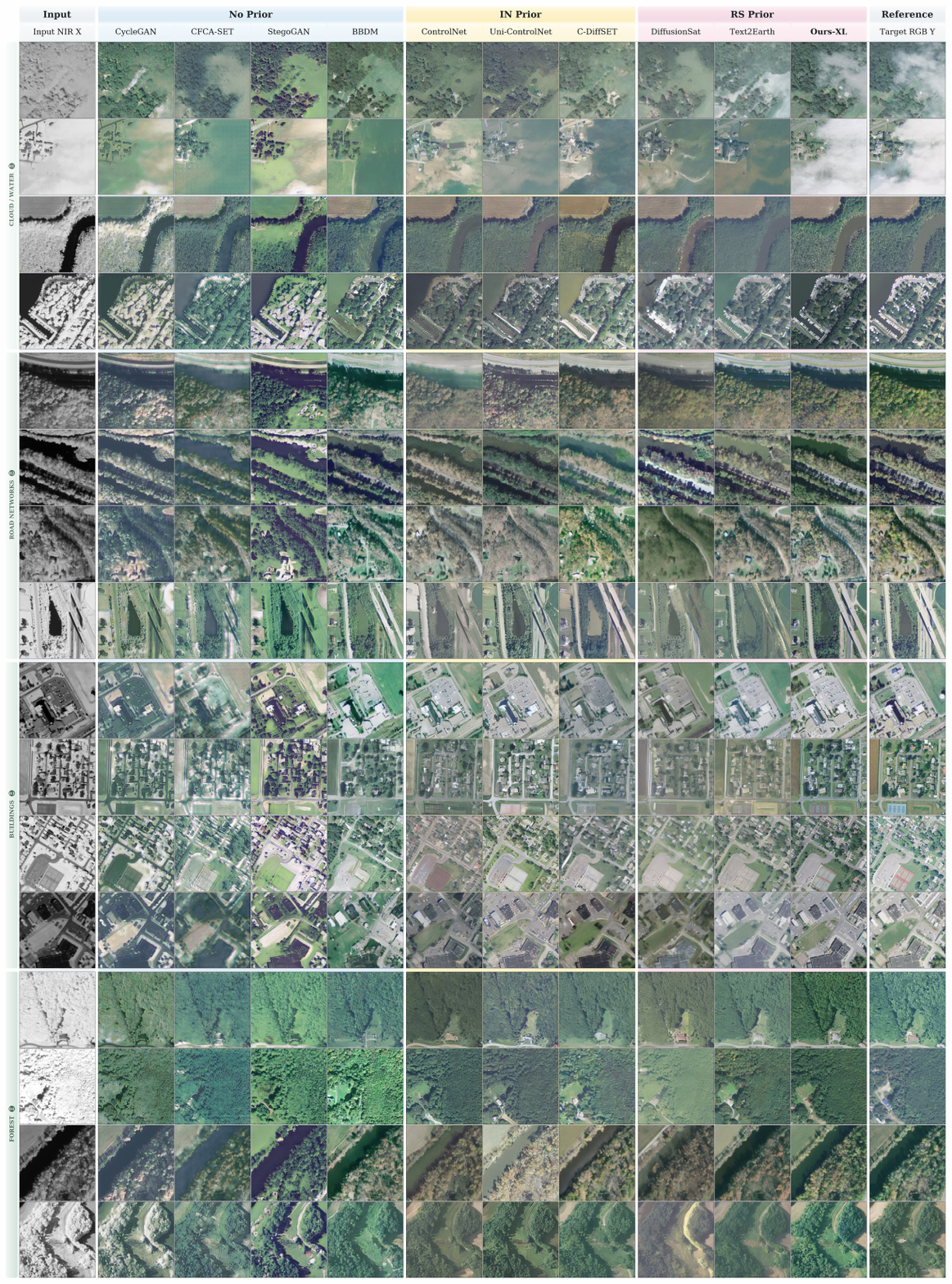}
\caption{Qualitative comparisons on the Chesapeake dataset.}
\label{fig:result_chesapeake}
\end{figure*}

\clearpage

\bibliography{references}